\pdfoutput=1
\documentclass[dvipsnames,format=sigconf,authorversion,nonacm]{acmart}

\usepackage{algorithm}
\usepackage{algpseudocode}
\algtext*{EndWhile}
\algtext*{EndIf}
\algtext*{EndFor}
\algtext*{EndProcedure}

\usepackage{subcaption}
\usepackage{booktabs}
\usepackage{tabularx}
\usepackage{todonotes}
\usepackage[export]{adjustbox}
\usepackage{multirow}

\usepackage{mathtools}
\usepackage{bm}

\newcommand{\FOS}{\mathcal{F}}
\newcommand{\IND}{\mathcal{I}}
\newcommand{\VAR}{\bm{X}}
\newcommand{\SUB}{\mathbb{I}}
\newcommand{\SEL}{\mathcal{S}}
\newcommand{\POP}{\mathcal{P}}

\newcommand{\CLI}{\mathcal{C}}

\newcommand{\bftab}{\fontseries{b}\selectfont}

\usepackage{tikz}
\usetikzlibrary{graphs,calc,arrows}

\newif\ifedges
\edgestrue

\usepackage{pifont}

\usepackage{float}
\usepackage{dblfloatfix}

\AtBeginDocument{%
  \providecommand\BibTeX{{%
    \normalfont B\kern-0.5em{\scshape i\kern-0.25em b}\kern-0.8em\TeX}}}

\setcopyright{acmcopyright}
\copyrightyear{2024}
\acmYear{2024}
\acmDOI{X}

\acmConference[GECCO '24]{GECCO '24:  Genetic and Evolutionary Computation Conference}{July 14--18, 2024}{Melbourne, Australia}
\acmBooktitle{GECCO '24: Genetic and Evolutionary Computation Conference, July 14--18, 2024, Melbourne, Australia}
\acmPrice{15.00}
\acmISBN{XXXXXXXXXX}

\begin{document}

\emergencystretch 3em
\setlength{\dbltextfloatsep}{7pt}
\setlength{\textfloatsep}{7pt}

\title[Fitness-based Linkage Learning and Maximum-Clique Conditional Linkage Modelling for GBO with RV-GOMEA]{Fitness-based Linkage Learning and Maximum-Clique Conditional Linkage Modelling for Gray-box Optimization\\with RV-GOMEA}





\author{Georgios Andreadis}
\affiliation{%
  \institution{Leiden University Medical Center}
  \city{Leiden}
  \country{The Netherlands}
}
\email{G.Andreadis@lumc.nl}

\author{Tanja Alderliesten}
\affiliation{%
  \institution{Leiden University Medical Center}
  \city{Leiden}
  \country{The Netherlands}
}
\email{T.Alderliesten@lumc.nl}

\author{Peter A.N. Bosman}
\affiliation{%
  \institution{Centrum Wiskunde \& Informatica}
  \city{Amsterdam}
  \country{The Netherlands}
}
\email{Peter.Bosman@cwi.nl}

\renewcommand{\shortauthors}{Andreadis et al.}

\begin{abstract}

For many real-world optimization problems it is possible to perform partial evaluations, meaning that the impact of changing a few variables on a solution's fitness can be computed very efficiently. It has been shown that such partial evaluations can be excellently leveraged by the Real-Valued GOMEA (RV-GOMEA) that uses a linkage model to capture dependencies between problem variables. Recently, conditional linkage models were introduced for RV-GOMEA, expanding its state-of-the-art performance even to problems with overlapping dependencies. However, that work assumed that the dependency structure is known a priori. Fitness-based linkage learning techniques have previously been used to detect dependencies during optimization, but only for non-conditional linkage models. In this work, we combine fitness-based linkage learning and conditional linkage modelling in RV-GOMEA. In addition, we propose a new way to model overlapping dependencies in conditional linkage models to maximize the joint sampling of fully interdependent groups of variables. We compare the resulting novel variant of RV-GOMEA to other variants of RV-GOMEA and VkD-CMA on 12 problems with varying degree of overlapping dependencies. We find that the new RV-GOMEA not only performs best on most problems, also the overhead of learning the conditional linkage models during optimization is often negligible.

\end{abstract}

\ccsdesc[500]{Mathematics of computing~Evolutionary algorithms}

\keywords{Gray-box optimization, GOMEA, linkage modeling}

\maketitle

\newcommand{\rebfig}{3.65cm}
\newcommand{\rebspace}{\hspace*{0.1cm}}
\newcommand{\rebgap}{0.3cm}

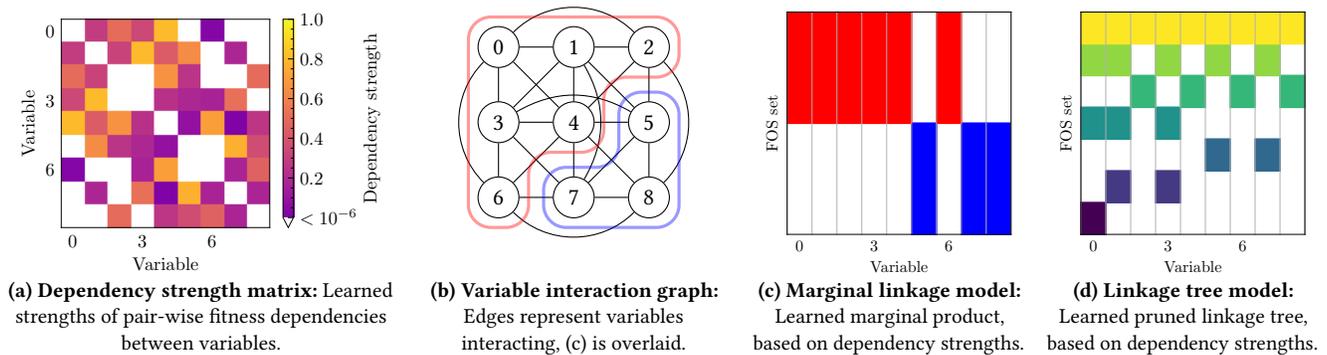
\begin{figure*}
    \setlength{\belowcaptionskip}{-3pt}
    \centering
    \subfloat[\label{fig:reb-grid-example:dsm}\centering \textbf{Dependency strength matrix:} {\normalfont Learned strengths of pair-wise fitness dependencies between variables.}]{\rebspace\includegraphics[height=\rebfig,valign=c]{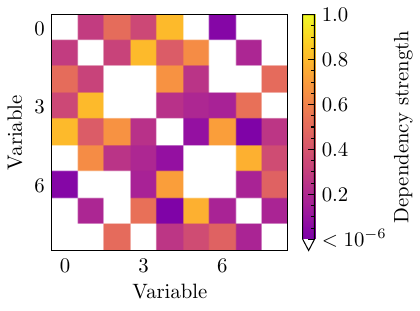}\rebspace}%
    \hspace{\rebgap}%
    \subfloat[\label{fig:reb-grid-example:vig}\centering \textbf{Variable interaction graph:} {\normalfont Edges represent variables interacting, (c) is overlaid.}]{
    \rebspace
        \begin{tikzpicture}[->]

        \newcommand{\bba}{-0.9}
        \newcommand{\bbb}{2.9}
        
        \draw[opacity=0] (\bba,\bba) -- (\bba,\bbb) -- (\bbb,\bbb) -- (\bbb,\bba) -- cycle;

        \edgestrue
        \begin{scope}[every node/.style={circle,draw,fill=white}]
    \node (0) at (0,2) {0};
    \node (1) at (1,2) {1};
    \node (2) at (2,2) {2};
    \node (3) at (0,1) {3};
    \node (4) at (1,1) {4};
    \node (5) at (2,1) {5};
    \node (6) at (0,0) {6};
    \node (7) at (1,0) {7};
    \node (8) at (2,0) {8};
\end{scope}

\definecolor{lightlightgray}{rgb}{0.85,0.85,0.85}

\ifedges
  \newcommand{\edgecol}{black}
\else
  \newcommand{\edgecol}{lightlightgray}
\fi

\begin{scope}[every node/.style={fill=white,circle},
              every edge/.style={\edgecol}]
    \draw[-,\edgecol] (0) -- (1);
    \draw[-,\edgecol] (0) -- (4);
    \draw[-,\edgecol] (0) -- (3);
    \draw[-,\edgecol] (1) -- (3);
    \draw[-,\edgecol] (1) -- (4);
    \draw[-,\edgecol] (1) -- (5);
    \draw[-,\edgecol] (1) -- (2);
    \draw[-,\edgecol] (2) -- (4);
    \draw[-,\edgecol] (2) -- (5);
    \draw[-,\edgecol] (3) -- (4);
    \draw[-,\edgecol] (3) -- (7);
    \draw[-,\edgecol] (3) -- (6);
    \draw[-,\edgecol] (4) -- (6);
    \draw[-,\edgecol] (4) -- (7);
    \draw[-,\edgecol] (4) -- (8);
    \draw[-,\edgecol] (4) -- (5);
    \draw[-,\edgecol] (5) -- (7);
    \draw[-,\edgecol] (5) -- (8);
    \draw[-,\edgecol] (6) -- (7);
    \draw[-,\edgecol] (7) -- (8);

    \draw[-,\edgecol]  (0) to [out=45,in=135] (2);
    \draw[-,\edgecol]  (3) to [out=30,in=150] (5);
    \draw[-,\edgecol]  (6) to [out=315,in=225] (8);
    \draw[-,\edgecol]  (0) to [out=225,in=135] (6);
    \draw[-,\edgecol]  (2) to [out=315,in=45] (8);
    \draw[-,\edgecol]  (1) to [out=300,in=60] (7);
\end{scope}
        
        \draw[red, very thick, rounded corners=3mm, opacity=0.4] (-0.4,2.4) -- (2.4,2.4) -- (2.4,1.6) -- (1.4,1.6) -- (1.4,0.6) -- (0.4,0.6) -- (0.4,-0.4) -- (-0.4,-0.4) -- cycle;
        
        \draw[blue, very thick, rounded corners=3mm, opacity=0.4] (0.6,-0.4) -- (2.4,-0.4) -- (2.4,1.4) -- (1.6,1.4) -- (1.6,0.4) -- (0.6,0.4) -- cycle;
        
        \end{tikzpicture}
        \vspace{0.2cm}
        \rebspace
    }%
    \hspace{\rebgap}%
    \subfloat[\label{fig:reb-grid-example:fos-mp}\centering \textbf{Marginal linkage model:} {\normalfont Learned marginal product, based on dependency strengths.}]{\rebspace\includegraphics[height=\rebfig,valign=c]{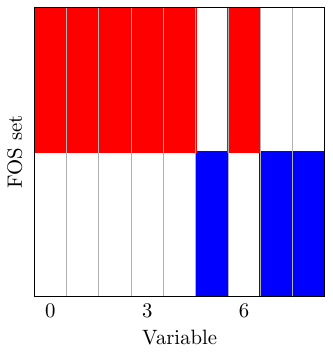}\rebspace}%
    \hspace{\rebgap}%
    \subfloat[\label{fig:reb-grid-example:fos-lt}\centering \textbf{Linkage tree model:} {\normalfont Learned pruned linkage tree, based on dependency strengths.}]{\rebspace\includegraphics[height=\rebfig,valign=c]{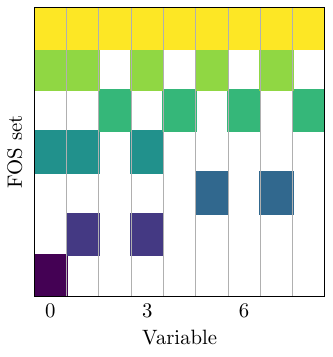}\rebspace}%
    \caption{A case study of fitness-based linkage learning: the \textit{REBGrid} problem described in Section~\ref{sec:experiments:problems}, with $\ell=9$.}
    \label{fig:reb-grid-example}
\end{figure*}

\vspace{0.2cm}

\section{Introduction}

Optimizing real-world problems using a Black-Box Optimization (BBO) perspective is generally applicable, but this perspective ignores problem-specific knowledge, despite it being available in many real-world optimization settings~\cite{Deb2016,Andreadis2023,Dickhoff2022}.
In a Gray-Box Optimization (GBO) setting, any such knowledge is provided to the optimization method, which can take many forms.
It has been demonstrated that Evolutionary Algorithms (EAs) can be adjusted to exploit such knowledge to improve performance, using techniques such as partition crossover operators~\cite{Carvalho2019}, problem-specific variation operators~\cite{Deb2016}, and partial evaluations~\cite{Bouter2017c}.
The latter assumes that it is possible to efficiently recalculate the objective value of a solution when a subset of its variables is modified.

For various real-world optimization problems, such as medical deformable image registration~\cite{Andreadis2023} and internal radiation treatment planning for cancer~\cite{Dickhoff2022}, the evaluation of solutions can be decomposed into such partial evaluations.
Despite their decomposability, however, these problems remain difficult to solve, often having characteristics such as non-smoothness, non-separability, and multi-modality.
The Real-Valued Gene-pool Optimal Mixing Evolutionary Algorithm (RV-GOMEA) has shown excellent performance on these real-world GBO problems~\cite{Andreadis2023,Dickhoff2022}, as well as on benchmark GBO problems~\cite{Bouter2017c,Bouter2017b}.
On optimization problems with strong overlapping dependencies, the recent introduction of conditional linkage models in RV-GOMEA has proven to be crucial to maintaining this performance improvement over a black-box approach~\cite{Bouter2020b}.

These conditional linkage models currently require the Variable Interaction Graph (VIG) to be provided, as dependencies between variables are core to conditional distribution sampling.
However, the interactions between problem variables may not be fully known in advance.
Even if a generic VIG formulation can be made, not all edges of this VIG may materialize as (strong) fitness dependencies during optimization.
In these cases, it may prove beneficial to empirically learn the VIG of a problem during optimization, as has been proposed in discrete optimization~\cite{Przewozniczek2022}.
On top of potentially speeding up optimization, this could unlock problem-specific, structural insights, which could be transferred to other problem instances.
For these instances, linkage then no longer needs to be detected during optimization, resulting in further speed-ups.
  
In this work, we bring the performance benefits of conditional linkage-models in RV-GOMEA~\cite{Bouter2020b} to GBO problems where the VIG is not known a priori.
To this end, we explore the use of fitness-based dependency tests to incrementally build a Dependency Strength Matrix (DSM), describing pairwise variable interactions.
These tests have recently shown promise within RV-GOMEA on separable problems~\cite{Olieman2021}, but have not yet been combined with conditional models.
Furthermore, we introduce a novel linkage model that models fully interdependent groups of variables jointly, maximizing the exploitation of sub-functions in the variation operator of GOMEA.
In experiments, we compare the performance of RV-GOMEA using these models to VkD-CMA~\cite{Akimoto2016,Akimoto2016a}, which is a state-of-the-art optimization method for problems with overlapping dependencies.
Although VkD-CMA is not able to leverage partial evaluations, it has a dimensionality-reduction mechanism capable of tackling high-dimensional problems with overlapping dependencies.

The remainder of this paper is structured as follows: First, we introduce the problem setting in Section~\ref{sec:gbo} and RV-GOMEA in Section~\ref{sec:rv-gomea}.
In Section~\ref{sec:learning}, we describe how linkage models can be learned using fitness-based dependency tests.
We evaluate and compare the proposed linkage models in RV-GOMEA to existing linkage models and VkD-CMA, in Sections~\ref{sec:experiments} and~\ref{sec:results}.
Finally, we discuss our results in Section~\ref{sec:discussion} and conclude in Section~\ref{sec:conclusions}.

\section{Gray-Box Optimization}
\label{sec:gbo}

We consider an objective function $f(\bm{x}) : \mathbb{R}^\ell \to \mathbb{R}$ which is to be minimized.
We refer to the variables of the problem as $\VAR = \{ \VAR_{0}, \dots, \VAR_{\ell - 1} \}$, indexed by the set of indices $\IND = [0, \dots, \ell - 1]$.
A realization of $\VAR$, i.e., a solution, is denoted by $\bm{x} = \{ \bm{x}_0, \dots, \bm{x}_{\ell - 1} \}$.

The GBO problems considered in this work allow for partial evaluations, which is the efficient computation of the objective value of a solution after a subset of its variables has been modified~\cite{Bouter2018}.
Let $\bm{Y} \subseteq \IND$ denote the indices of this subset of variables, and $\bm{x}_{\bm{Y}}$ the corresponding variables.
The objective function $f(\bm{x})$ is then composed of $q$ sub-functions $\bm{F} = \{ f_0, \dots, f_{q-1} \}$.
Each sub-function $f_j(\bm{x}_{\SUB_j}) \in \bm{F}$ takes variables $\bm{x}_i$ with $i \in \SUB_j \subseteq \IND$, where sub-function index sets $\SUB = \{ \SUB_0, \dots, \SUB_{q-1} \}$ are given by the problem definition.
These sub-functions are considered not to be separable themselves, and are therefore treated as a black box.

As such, a GBO objective function can be expressed as follows:
\begin{equation}
    f(\bm{x}) = f_0(\bm{x}_{\SUB_0}) \oplus f_1(\bm{x}_{\SUB_1}) \oplus \dots \oplus f_{q-1}(\bm{x}_{\SUB_{q-1}}) ,
\end{equation}
with $\oplus$ denoting a binary operator with a known inverse operator~$\ominus$ (such as addition or multiplication).

When one variable $x_u$ is changed, all sub-functions $f_j(\bm{x}_{\IND_j})$ for which $u \in \IND_j$ holds, need to be evaluated.
We consider the computational complexity of a sub-function $f_j$ to be approximately proportional to the number of involved variables $|\IND_j|$.
Therefore, the cost of a partial evaluation for $f_j$ is calculated as $|\IND_j| / |\IND|$ in this setting.
This deviates from a previous definition~\cite{Bouter2020b} which does not take sub-function size into account.
The definition used in this work portrays problems with heterogeneous sub-function index set sizes and strongly overlapping dependencies more accurately.

In this gray-box structure, we consider two different variables $\bm{x}_u$ and $\bm{x}_v$ with indices $u, v \in \IND$ to be directly dependent when there exists a sub-function $f_j$ with $\{u, v\} \subseteq \IND_j$.
This dependency is denoted by $\bm{x}_u \leftrightarrow \bm{x}_v$.
Variables $\bm{x}_u$ and $\bm{x}_v$ are considered indirectly dependent when there exists a set $\{u, \dots, v\} \subseteq \IND$ for which $\bm{x}_u \leftrightarrow \dots \leftrightarrow \bm{x}_v$ holds, but not $\bm{x}_u \leftrightarrow \bm{x}_v$.
Given the dependencies in a problem, a so-called Variable Interaction Graph (VIG)~\cite{Tinos2015} can be constructed.
The VIG of a problem is an undirected graph $\bm{G} = (\bm{V}, \bm{E})$, with each vertex $v \in \bm{V}$ corresponding to the variable $\bm{x}_v$ and an edge $(u, v) \in \bm{E}$ exists for every pair of directly dependent variables $\bm{x}_u$ and $\bm{x}_v$.
Indirect dependencies can be translated to this graph representation as follows: if $\bm{x}_u$ and $\bm{x}_v$ are indirectly independent, there exists a path between $u$ and $v$, but not an edge $(u, v)$.
As illustration, the VIG of the \textit{REBGrid} problem (defined in Section~\ref{sec:experiments:problems}) is depicted in Figure~\ref{fig:reb-grid-example:vig}.


\section{Real-Valued Gene-Pool Optimal Mixing Evolutionary Algorithm}
\label{sec:rv-gomea}

In this section, we present an overview of RV-GOMEA.
The algorithm is described in more detail in~\cite{Bouter2017c}.


\subsection{Family of Subsets}

The dependencies between variables $\VAR_i$ are modelled explicitly by a linkage model in RV-GOMEA.
This linkage model is described by a Family of Subsets (FOS) $\FOS = \{ \FOS_0, \FOS_1, \dots, \FOS_{m-1} \}$, with $\FOS_i \subseteq \IND$.
The elements $\FOS_i$ of this FOS each represent a group of variables which are deemed to be jointly dependent.

Two generic FOS linkage models can be defined regardless of the problem structure: a univariate FOS $\FOS = \{ \{0\}, \{1\}, \dots, \{\ell - 1\} \}$, which models all variables to be independent, and a full FOS $\FOS~=~\{ \{ 0, \dots, \ell - 1 \} \}$, which models all variables as jointly dependent.
More generally, we can define a marginal product FOS model, as a set of disjoint sets of dependent variables, i.e., $\FOS_i \cap \FOS_j = \emptyset$ for all $\FOS_i, \FOS_j \in \FOS$ where $i \neq j$.
An example is visualized in Figure~\ref{fig:reb-grid-example:fos-mp}.

As a generic solution that can capture different orders of dependency at once, the linkage tree FOS model has been introduced~\cite{Thierens2011,Bouter2017c}.
To construct such an FOS, all singleton elements are first included.
Then, elements are recursively merged into larger elements $\FOS_k$, for which the following holds: $\FOS_i, \FOS_j \in \FOS$ exist ($i \neq j \neq k$) such that $\FOS_i \cap \FOS_j = \emptyset$ and $\FOS_i \cup \FOS_j = \FOS_k$.
To decide which $\FOS_i, \FOS_j$ to merge, the Unweighted Pair Grouping Method with Arithmetic-mean (UPGMA)~\cite{Gronau2007} is commonly used.
This method is supplied with dependency strength information, which can be provided \textit{a priori} or can be discovered during optimization, for instance by measuring mutual information in the population, although this technique has proven to be ineffective in continuous optimization~\cite{Olieman2021}.

A linkage tree FOS can be restricted in size from two directions.
From the root of the tree, its elements can be bounded to a maximum element size, which is then referred to as a bounded linkage tree~\cite{Bouter2017c}.
From the leaves of the tree upward, the model can exclude smaller elements which are better solely modelled jointly, referred to as a pruned linkage tree~\cite{Olieman2021}.
This pruning is iteratively performed until there exist no $\FOS_i,\FOS_j,\FOS_k$ ($i \neq j \neq k$) for which the following two conditions hold: $\FOS_i \cup \FOS_j = \FOS_k$, and all variables in $\FOS_k$ are jointly dependent.
If a group of elements exists for which this holds, only the largest element $\FOS_k$ is preserved, since this element is considered to model the joint dependencies sufficiently.
An example of a pruned linkage tree FOS is visualized in Figure~\ref{fig:reb-grid-example:fos-lt}.

\vspace{-0.2cm}


\begin{algorithm}[H]
\small
\caption{RV-GOMEA}\label{alg:rv-gomea}

\begin{algorithmic}[1]

\Procedure{RV-GOMEA}{$f, n, \tau$}
  \State $\POP \gets \texttt{InitializeAndEvaluatePopulation}(f, n)$
  \State $\FOS \gets \texttt{InitializeLinkageModel}(f)$
  \While{$\neg \texttt{TerminationCriterionSatisfied}()$}
    \State $\POP_0 \gets \bm{x}^{\texttt{elitist}}$
    
    \For{$\FOS_i \in \FOS$} \Comment{In random order}
        \State $\SEL \gets \lfloor \tau n \rfloor \text{ best in } \POP$
        \State $P(\{ \bm{X}_u : u \in \FOS_i \}) \gets \texttt{MaxLikelihoodEstimate}(\SEL)$

        \For{$\bm{x} \in \POP_{1 \dots n-1}$}
            \State $\texttt{GenepoolOptimalMixing}(f, \bm{x}, \FOS_i)$
        \EndFor
        
        \State $\texttt{AdaptiveVarianceScaling}(\FOS_i)$
    \EndFor
    
    \For{$\bm{x} \in \POP_{1 \dots n^{\text{AMS}}}$}
        \State $\texttt{AnticipatedMeanShift}(\bm{x})$
    \EndFor
    
    \For{$\bm{x} \in \POP_{1 \dots n-1}$}
        \If{$\text{NIS}(\bm{x}) > \text{NIS}^{\text{MAX}}$}
            \State $\texttt{ForcedImprovement}(\bm{x})$
        \EndIf
    \EndFor
  \EndWhile
\EndProcedure

\end{algorithmic}

\end{algorithm}
\vspace{-0.3cm}

\vspace{-0.4cm}

\begin{algorithm}[H]
\small
\caption{Gene-pool Optimal Mixing}\label{alg:gom}

\begin{algorithmic}[1]

\Procedure{GenepoolOptimalMixing}{$f, \bm{x}, \FOS_i$}
  \State $\bm{a} \gets \bm{x}_{\FOS_i}$ \Comment{Retain original}

  \State $\bm{x}_{\FOS_i} \gets P(\{ \bm{X}_u : u \in \FOS_i \})$
  \State $f_{\bm{x}}' \gets \texttt{Evaluate}(\bm{x}, f_{\bm{x}}, \FOS_i)$ \Comment{Partially evaluate, if possible}
  
  \State $\textbf{if } f_{\bm{x}}' < f_{\bm{x}} \textbf{ or } \mathcal{U} < p^{\texttt{accept}} \textbf{ then } f_{\bm{x}} \gets f_{\bm{x}}'$

  \State $\textbf{else } \bm{x}_{\FOS_i} \gets \bm{a}$
\EndProcedure

\end{algorithmic}

\end{algorithm}


\vspace{-0.4cm}

\subsection{Gene-pool Optimal Mixing}
\label{sec:rv-gomea:gom}

The variation operator of RV-GOMEA is a mechanism called Gene-pool Optimal Mixing (GOM).
In GOM, variation is applied to solutions based on the dependencies encoded in the linkage model $\FOS$.
GOM is applied at each generation, to each solution in the population, and for each FOS element $\FOS_i$.
At each application of GOM, new values are sampled for the variables contained in $\FOS_i$, inserted into a parent solution, and evaluated (by partial evaluation).
If this modification improves the parent solution's objective value, it is accepted.
Otherwise, it may still be accepted with probability $p^{\texttt{accept}}$, or else be rejected, in which case the parent's genotype is restored.

The GOM operator uses a sampling model to sample new variables.
This work uses both the AMaLGaM sampling model~\cite{Bosman2013} and the conditional RV-GOMEA sampling model~\cite{Bouter2020b} (described in Section~\ref{sec:rv-gomea:conditional}).
Previous work has shown that also models from other EAs, such as CMA-ES~\cite{Hansen2003}, can be integrated in GOM~\cite{Bouter2020}.

The AMalGaM-based sampling model is based on a Gaussian distribution that is estimated using maximum likelihood based on selected solutions $\SEL$ and is adapted to prevent premature convergence as well as better alignment with the direction of largest improvement in the landscape.
These adaptation details are explained in further detail in~\cite{Bouter2017c}.
Here, we include pseudocode of RV-GOMEA, in Algorithm~\ref{alg:rv-gomea}, and the GOM operator in particular, in Algorithm~\ref{alg:gom}.
This definition of RV-GOMEA follows the definition in~\cite{Bouter2020b} by performing GOM at each FOS element iteration, as opposed to each generation~\cite{Bouter2017c}.

\subsection{Conditional Gene-pool Optimal Mixing}
\label{sec:rv-gomea:conditional}

The above described marginal product and linkage tree FOS models have shown to lead to good performance of RV-GOMEA on problems without overlapping dependencies~\cite{Bouter2017c}.
However, many real-world GBO problems have overlapping dependency structures.
Previous work has indicated that problems with increasing dependency strength are increasingly difficult to optimize with such linkage models~\cite{Bouter2020b}, as local GOM steps on certain structures can break linkage with other, overlapping structures.
Directly modeling the network of dependencies between variables may be key to more efficiently optimizing such overlapping structures.

\subsubsection{Bayesian and Markov network models}
Dependency networks such as Bayesian networks have been included in Estimation of Distribution Algorithms (EDAs) before, for both discrete~\cite{Pelikan1999,Pelikan2005,Echegoyen2007} and real-valued problems~\cite{Ahn2004,Bosman2009}. 
Bayesian network-based EDAs such as the Bayesian Optimization Algorithm (BOA)~\cite{Pelikan2005,Pelikan1999} learn a network from a selected set of solutions, using a quality metric such as Bayesian Information Criterion (BIC) score.
Similarly, Markov networks have also been used in EDAs~\cite{Shakya2012}.
Examples for discrete optimization include the Distribution Estimation Using Markov networks (DEUM) algorithm~\cite{Shakya2010}, the Markovianity-based Optimization Algorithm (MOA)~\cite{Shakya2008}, and the Markovian Learning Estimation of Distribution Algorithm (MARLEDA)~\cite{Alden2016}.
The Gaussian Markov Random Field EDA (GMRF-EDA)~\cite{Karshenas2012} is an example of a real-valued EDA using a factorized Gaussian Markov network.
However, it is focused on modeling disjoint dependency structures.

In RV-GOMEA, a conditional linkage model has been recently introduced which derives a Gaussian Markov Field (GMF) from the VIG of a problem and estimates a conditional distribution based on a GMF traversal and factorization.
This strategy has proven to be effective when optimizing problems with overlapping dependency structures~\cite{Bouter2020b}.
Multiple variants of this model exist, featuring different factorizations of the GMF and being applied at different levels of GOM optimization.
These variants are described below.

\subsubsection{GMF factorization methods}
Two factorization methods have been proposed within RV-GOMEA.
Both conduct a breadth-first traversal of the GMF to obtain a factorized conditional distribution.
The first, \textit{UCond}, effectively translates the GMF directly to a Bayesian network, modelling $\ell$ factors that represent a conditional density function of one variable, given a set of parent variables it depends on.
Dependent variables encountered during traversal are recorded for later conditional sampling, in the direction that traversal occurred.
The second, \textit{MCond}, conducts a clique search during traversal, merging groups of unvisited variables which are all connected by edges in the VIG and all depend on the same set of dependent variables into clique elements~\cite{Bouter2020b}.
This clique search effectively finds a marginal product partitioning of the VIG, with each found clique being modelled jointly afterward.
These cliques are often not maximal, since their members need to be fully connected and conditioned on the same dependent variables.
Of note: the conditional distributions defined by the MCond and UCond factorizations are probabilistically identical, i.e., they encode the same conditional distribution over all variables, but different FOS elements are identified for GOM.

An example MCond factorization is visualized in Figure~\ref{fig:reb-grid-vig:old}, next to the derived factorized conditional distribution, in Figure~\ref{fig:reb-grid-vig:old:network}.
This factorization encodes the following probability distribution:
\begin{align*}
    P(\bm{X}) 
    =~&P \left( \bm{X}_{\{ 0, 1, 2, 4 \}} \right) 
    * P \left( \bm{X}_{\{ 3 \}} \;\middle|\; \bm{X}_{\{ 0, 1, 4 \}} \right) \\
    *~&P \left( \bm{X}_{\{ 5 \}} \;\middle|\; \bm{X}_{\{ 1,2,3,4 \}} \right)
    * P \left( \bm{X}_{\{ 6 \}} \;\middle|\; \bm{X}_{\{ 0,3,4,7 \}} \right) \\
    *~&P \left( \bm{X}_{\{ 7 \}} \;\middle|\; \bm{X}_{\{ 1,3,4,5 \}} \right) 
    * P \left( \bm{X}_{\{ 8 \}} \;\middle|\; \bm{X}_{\{ 2,4,5,6,7 \}} \right)
\end{align*}

\subsubsection{GOM optimization levels}
The derived factorized conditional distribution can be leveraged at different levels.
At the Factorized GOM (FG) level, both conditional FOS elements and non-conditional FOS elements, pertaining to respectively conditional dependencies and joint dependencies are identified.
In the factorized distribution above, each factor corresponds to an FOS element.
This results in $\ell$ conditional FOS elements in the UCond factorization, and at most $\ell$ elements in the MCond factorization.
GOM is applied to each FOS element separately.
For a non-conditional FOS element, GOM is equivalent to standard GOM as outlined in Section~\ref{sec:rv-gomea:gom}.
For a conditional element (i.e., conditional GOM), GOM is essentially the same, but samples from the conditional distribution where the values for the variables being conditioned on are taken from the solution that is undergoing GOM.

At the Generational GOM (GG) level, the entire factorized conditional distribution, consisting of all variables, is modelled as one conditional FOS element, and sampled in one forward sampling step.
Although this requires a full solution evaluation, this step can repair linkages that GOM steps on smaller FOS elements may have broken, because each GOM step ends with a separate selection step.
For this reason, a Hybrid GOM (HG) mode has been proposed, which combines both levels.
This performed best on various optimization problems~\cite{Bouter2020b}.

In previous work, the order in which FOS elements of the FG and GG levels are optimized within one generation of HG was random.
In this work, we have adapted this order to always perform GOM at the GG level first, followed by GOM on all smaller FG elements, in random order.
This is based on preliminary experiments, showing a slight improvement in performance across the problems considered in this paper if the full GG element is used first.

\newcommand{\viggap}{0.1cm}

\newcommand{\vbba}{-0.8}
\newcommand{\vbbb}{2.8}

\newcommand{\scale}{0.8}
\newcommand{\sep}{0.4mm}
\newcommand{\op}{0.7}

\edgesfalse

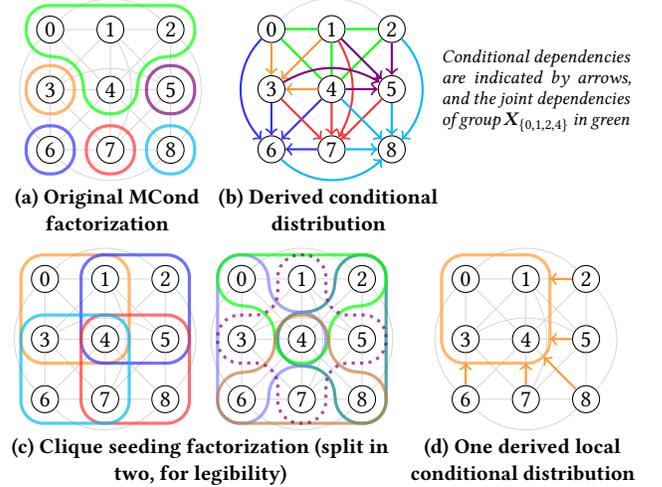
\begin{figure}
    \centering
    \subfloat[\label{fig:reb-grid-vig:old}\centering Original MCond factorization]{
    \begin{tikzpicture}[->,scale=\scale, every node/.style={scale=\scale},inner sep=\sep]

    
    \begin{scope}[every node/.style={circle,draw,fill=white}]
    \node (0) at (0,2) {0};
    \node (1) at (1,2) {1};
    \node (2) at (2,2) {2};
    \node (3) at (0,1) {3};
    \node (4) at (1,1) {4};
    \node (5) at (2,1) {5};
    \node (6) at (0,0) {6};
    \node (7) at (1,0) {7};
    \node (8) at (2,0) {8};
\end{scope}

\definecolor{lightlightgray}{rgb}{0.85,0.85,0.85}

\ifedges
  \newcommand{\edgecol}{black}
\else
  \newcommand{\edgecol}{lightlightgray}
\fi

\begin{scope}[every node/.style={fill=white,circle},
              every edge/.style={\edgecol}]
    \draw[-,\edgecol] (0) -- (1);
    \draw[-,\edgecol] (0) -- (4);
    \draw[-,\edgecol] (0) -- (3);
    \draw[-,\edgecol] (1) -- (3);
    \draw[-,\edgecol] (1) -- (4);
    \draw[-,\edgecol] (1) -- (5);
    \draw[-,\edgecol] (1) -- (2);
    \draw[-,\edgecol] (2) -- (4);
    \draw[-,\edgecol] (2) -- (5);
    \draw[-,\edgecol] (3) -- (4);
    \draw[-,\edgecol] (3) -- (7);
    \draw[-,\edgecol] (3) -- (6);
    \draw[-,\edgecol] (4) -- (6);
    \draw[-,\edgecol] (4) -- (7);
    \draw[-,\edgecol] (4) -- (8);
    \draw[-,\edgecol] (4) -- (5);
    \draw[-,\edgecol] (5) -- (7);
    \draw[-,\edgecol] (5) -- (8);
    \draw[-,\edgecol] (6) -- (7);
    \draw[-,\edgecol] (7) -- (8);

    \draw[-,\edgecol]  (0) to [out=45,in=135] (2);
    \draw[-,\edgecol]  (3) to [out=30,in=150] (5);
    \draw[-,\edgecol]  (6) to [out=315,in=225] (8);
    \draw[-,\edgecol]  (0) to [out=225,in=135] (6);
    \draw[-,\edgecol]  (2) to [out=315,in=45] (8);
    \draw[-,\edgecol]  (1) to [out=300,in=60] (7);
\end{scope}

    \draw[orange!80, very thick, rounded corners=3mm, opacity=\op] (-0.4,1.4) -- (0.4,1.4) -- (0.4,0.6) -- (-0.4,0.6) -- cycle;

    \draw[blue!80, very thick, rounded corners=3mm, opacity=\op] (-0.4,0.4) -- (0.4,0.4) -- (0.4,-0.4) -- (-0.4,-0.4) -- cycle;
    
    \draw[red!80, very thick, rounded corners=3mm, opacity=\op] (0.6,0.4) -- (1.4,0.4) -- (1.4,-0.4) -- (0.6,-0.4) -- cycle;

    \draw[cyan!80, very thick, rounded corners=3mm, opacity=\op] (1.6,0.4) -- (2.4,0.4) -- (2.4,-0.4) -- (1.6,-0.4) -- cycle;
    
    \draw[violet, very thick, rounded corners=3mm, opacity=\op] (1.6,1.4) -- (2.4,1.4) -- (2.4,0.6) -- (1.6,0.6) -- cycle;
    
    \draw[green, very thick, rounded corners=3mm, opacity=\op] (-0.4,2.4) -- (2.4,2.4) -- (2.4,1.6) -- (1.5,1.5) -- (1.4,0.6) -- (0.6,0.6) -- (0.5,1.5) -- (-0.4,1.6) -- cycle;

    \end{tikzpicture}
    }%
    \hspace{\viggap}%
    \subfloat[\label{fig:reb-grid-vig:old:network}\centering Derived conditional distribution]{\hspace*{0.1cm}
    \begin{tikzpicture}[->,scale=\scale, every node/.style={scale=\scale},inner sep=\sep]

    
    \begin{scope}[every node/.style={circle,draw,fill=white}]
    \node (0) at (0,2) {0};
    \node (1) at (1,2) {1};
    \node (2) at (2,2) {2};
    \node (3) at (0,1) {3};
    \node (4) at (1,1) {4};
    \node (5) at (2,1) {5};
    \node (6) at (0,0) {6};
    \node (7) at (1,0) {7};
    \node (8) at (2,0) {8};
\end{scope}

\definecolor{lightlightgray}{rgb}{0.85,0.85,0.85}

\ifedges
  \newcommand{\edgecol}{black}
\else
  \newcommand{\edgecol}{lightlightgray}
\fi

\begin{scope}[every node/.style={fill=white,circle},
              every edge/.style={\edgecol}]
    \draw[-,\edgecol] (0) -- (1);
    \draw[-,\edgecol] (0) -- (4);
    \draw[-,\edgecol] (0) -- (3);
    \draw[-,\edgecol] (1) -- (3);
    \draw[-,\edgecol] (1) -- (4);
    \draw[-,\edgecol] (1) -- (5);
    \draw[-,\edgecol] (1) -- (2);
    \draw[-,\edgecol] (2) -- (4);
    \draw[-,\edgecol] (2) -- (5);
    \draw[-,\edgecol] (3) -- (4);
    \draw[-,\edgecol] (3) -- (7);
    \draw[-,\edgecol] (3) -- (6);
    \draw[-,\edgecol] (4) -- (6);
    \draw[-,\edgecol] (4) -- (7);
    \draw[-,\edgecol] (4) -- (8);
    \draw[-,\edgecol] (4) -- (5);
    \draw[-,\edgecol] (5) -- (7);
    \draw[-,\edgecol] (5) -- (8);
    \draw[-,\edgecol] (6) -- (7);
    \draw[-,\edgecol] (7) -- (8);

    \draw[-,\edgecol]  (0) to [out=45,in=135] (2);
    \draw[-,\edgecol]  (3) to [out=30,in=150] (5);
    \draw[-,\edgecol]  (6) to [out=315,in=225] (8);
    \draw[-,\edgecol]  (0) to [out=225,in=135] (6);
    \draw[-,\edgecol]  (2) to [out=315,in=45] (8);
    \draw[-,\edgecol]  (1) to [out=300,in=60] (7);
\end{scope}

    \draw[-,green,thick] (0) -- (1);
    \draw[-,green,thick] (1) -- (2);
    \draw[-,green,thick] (0) -- (4);
    \draw[-,green,thick] (2) -- (4);
    \draw[-,green,thick] (1) -- (4);
    
    \draw[<-,orange!80,thick] (3) -- (0);
    \draw[<-,orange!80,thick] (3) -- (1);
    \draw[<-,orange!80,thick] (3) -- (4);
    
    \draw[<-,blue!80,thick] (6) to [out=135,in=225] (0);
    \draw[<-,blue!80,thick] (6) -- (3);
    \draw[<-,blue!80,thick] (6) -- (4);
    \draw[<-,blue!80,thick] (6) -- (7);

    \draw[<-,red!80,thick] (7) to [out=60,in=300] (1);
    \draw[<-,red!80,thick] (7) -- (3);
    \draw[<-,red!80,thick] (7) -- (4);
    \draw[<-,red!80,thick] (7) -- (5);
    
    \draw[<-,cyan!80,thick] (8) to [out=45,in=315] (2);
    \draw[<-,cyan!80,thick] (8) -- (4);
    \draw[<-,cyan!80,thick] (8) -- (5);
    \draw[<-,cyan!80,thick] (8) to [out=225,in=315] (6);
    \draw[<-,cyan!80,thick] (8) -- (7);
    
    \draw[<-,violet,thick] (5) -- (1);
    \draw[<-,violet,thick] (5) -- (2);
    \draw[<-,violet,thick] (5) to [out=150,in=30] (3);
    \draw[<-,violet,thick] (5) -- (4);
    


    \end{tikzpicture}%
    \hspace*{0.1cm}%
    }%
    \hspace{0.05cm}%
    %
    %
    %
    %
    %
    %
    %
    %
    %
    %
    \subfloat{\begin{minipage}{2.5cm}\footnotesize \textit{Conditional dependencies are indicated by arrows, and the joint dependencies of group $\bm{X}_{\{0, 1, 2, 4\}}$ in green}\vspace{0.75cm}\end{minipage}}
    \hspace{0.05cm}%
    \addtocounter{subfigure}{-1}
    \subfloat[\label{fig:reb-grid-vig:new}\centering Clique seeding factorization (split in two, for legibility)]{
    \begin{tikzpicture}[->,scale=\scale, every node/.style={scale=\scale},inner sep=\sep]
    
    
    \begin{scope}[every node/.style={circle,draw,fill=white}]
    \node (0) at (0,2) {0};
    \node (1) at (1,2) {1};
    \node (2) at (2,2) {2};
    \node (3) at (0,1) {3};
    \node (4) at (1,1) {4};
    \node (5) at (2,1) {5};
    \node (6) at (0,0) {6};
    \node (7) at (1,0) {7};
    \node (8) at (2,0) {8};
\end{scope}

\definecolor{lightlightgray}{rgb}{0.85,0.85,0.85}

\ifedges
  \newcommand{\edgecol}{black}
\else
  \newcommand{\edgecol}{lightlightgray}
\fi

\begin{scope}[every node/.style={fill=white,circle},
              every edge/.style={\edgecol}]
    \draw[-,\edgecol] (0) -- (1);
    \draw[-,\edgecol] (0) -- (4);
    \draw[-,\edgecol] (0) -- (3);
    \draw[-,\edgecol] (1) -- (3);
    \draw[-,\edgecol] (1) -- (4);
    \draw[-,\edgecol] (1) -- (5);
    \draw[-,\edgecol] (1) -- (2);
    \draw[-,\edgecol] (2) -- (4);
    \draw[-,\edgecol] (2) -- (5);
    \draw[-,\edgecol] (3) -- (4);
    \draw[-,\edgecol] (3) -- (7);
    \draw[-,\edgecol] (3) -- (6);
    \draw[-,\edgecol] (4) -- (6);
    \draw[-,\edgecol] (4) -- (7);
    \draw[-,\edgecol] (4) -- (8);
    \draw[-,\edgecol] (4) -- (5);
    \draw[-,\edgecol] (5) -- (7);
    \draw[-,\edgecol] (5) -- (8);
    \draw[-,\edgecol] (6) -- (7);
    \draw[-,\edgecol] (7) -- (8);

    \draw[-,\edgecol]  (0) to [out=45,in=135] (2);
    \draw[-,\edgecol]  (3) to [out=30,in=150] (5);
    \draw[-,\edgecol]  (6) to [out=315,in=225] (8);
    \draw[-,\edgecol]  (0) to [out=225,in=135] (6);
    \draw[-,\edgecol]  (2) to [out=315,in=45] (8);
    \draw[-,\edgecol]  (1) to [out=300,in=60] (7);
\end{scope}

    \draw[orange!80, very thick, rounded corners=3mm, opacity=\op] (-0.4,2.4) -- (1.4,2.4) -- (1.4,0.6) -- (-0.4,0.6) -- cycle;

    \draw[blue!80, very thick, rounded corners=3mm, opacity=\op] (0.6,2.4) -- (2.4,2.4) -- (2.4,0.6) -- (0.6,0.6) -- cycle;

    \draw[red!80, very thick, rounded corners=3mm, opacity=\op] (0.6,1.4) -- (2.4,1.4) -- (2.4,-0.4) -- (0.6,-0.4) -- cycle;

    \draw[cyan!80, very thick, rounded corners=3mm, opacity=\op] (-0.4,1.4) -- (1.4,1.4) -- (1.4,-0.4) -- (-0.4,-0.4) -- cycle;
    
    \end{tikzpicture}%
    \begin{tikzpicture}[->,scale=\scale, every node/.style={scale=\scale},inner sep=\sep]
    
    
    \begin{scope}[every node/.style={circle,draw,fill=white}]
    \node (0) at (0,2) {0};
    \node (1) at (1,2) {1};
    \node (2) at (2,2) {2};
    \node (3) at (0,1) {3};
    \node (4) at (1,1) {4};
    \node (5) at (2,1) {5};
    \node (6) at (0,0) {6};
    \node (7) at (1,0) {7};
    \node (8) at (2,0) {8};
\end{scope}

\definecolor{lightlightgray}{rgb}{0.85,0.85,0.85}

\ifedges
  \newcommand{\edgecol}{black}
\else
  \newcommand{\edgecol}{lightlightgray}
\fi

\begin{scope}[every node/.style={fill=white,circle},
              every edge/.style={\edgecol}]
    \draw[-,\edgecol] (0) -- (1);
    \draw[-,\edgecol] (0) -- (4);
    \draw[-,\edgecol] (0) -- (3);
    \draw[-,\edgecol] (1) -- (3);
    \draw[-,\edgecol] (1) -- (4);
    \draw[-,\edgecol] (1) -- (5);
    \draw[-,\edgecol] (1) -- (2);
    \draw[-,\edgecol] (2) -- (4);
    \draw[-,\edgecol] (2) -- (5);
    \draw[-,\edgecol] (3) -- (4);
    \draw[-,\edgecol] (3) -- (7);
    \draw[-,\edgecol] (3) -- (6);
    \draw[-,\edgecol] (4) -- (6);
    \draw[-,\edgecol] (4) -- (7);
    \draw[-,\edgecol] (4) -- (8);
    \draw[-,\edgecol] (4) -- (5);
    \draw[-,\edgecol] (5) -- (7);
    \draw[-,\edgecol] (5) -- (8);
    \draw[-,\edgecol] (6) -- (7);
    \draw[-,\edgecol] (7) -- (8);

    \draw[-,\edgecol]  (0) to [out=45,in=135] (2);
    \draw[-,\edgecol]  (3) to [out=30,in=150] (5);
    \draw[-,\edgecol]  (6) to [out=315,in=225] (8);
    \draw[-,\edgecol]  (0) to [out=225,in=135] (6);
    \draw[-,\edgecol]  (2) to [out=315,in=45] (8);
    \draw[-,\edgecol]  (1) to [out=300,in=60] (7);
\end{scope}

    \draw[teal, very thick, rounded corners=3mm, opacity=\op] (1.6,2.4) -- (2.4,2.4) -- (2.4,-0.4) -- (1.6,-0.4) -- (1.5,0.5) -- (0.6,0.6) -- (0.6,1.4) -- (1.5,1.5) -- cycle;

    \draw[blue!50, very thick, rounded corners=3mm, opacity=\op] (0.4,2.4) -- (-0.4,2.4) -- (-0.4,-0.4) -- (0.4,-0.4) -- (0.5,0.5) -- (1.4,0.6) -- (1.4,1.4) -- (0.5,1.5) -- cycle;

    \draw[green, very thick, rounded corners=3mm, opacity=\op] (-0.4,2.4) -- (2.4,2.4) -- (2.4,1.6) -- (1.5,1.5) -- (1.4,0.6) -- (0.6,0.6) -- (0.5,1.5) -- (-0.4,1.6) -- cycle;

    \draw[brown, very thick, rounded corners=3mm, opacity=\op] (-0.4,-0.4) -- (2.4,-0.4) -- (2.4,0.4) -- (1.5,0.5) -- (1.4,1.4) -- (0.6,1.4) -- (0.5,0.5) -- (-0.4,0.4) -- cycle;

    \draw[dotted, violet, very thick, rounded corners=3mm, opacity=\op] (-0.4,1.4) -- (0.6,1.4) -- (0.6,2.4) -- (1.4,2.4) -- (1.4,1.4) -- (2.4,1.4) -- (2.4,0.6) -- (1.4,0.6) -- (1.4,-0.4) -- (0.6,-0.4) -- (0.6,0.6) -- (-0.4,0.6) -- cycle;

    \end{tikzpicture}
    }%
    \hspace{\viggap}%
    \subfloat[\label{fig:reb-grid-vig:new:network}\centering One derived local conditional distribution]{\hspace*{0.105cm}
    \begin{tikzpicture}[->,scale=\scale, every node/.style={scale=\scale},inner sep=\sep]

    
    \begin{scope}[every node/.style={circle,draw,fill=white}]
    \node (0) at (0,2) {0};
    \node (1) at (1,2) {1};
    \node (2) at (2,2) {2};
    \node (3) at (0,1) {3};
    \node (4) at (1,1) {4};
    \node (5) at (2,1) {5};
    \node (6) at (0,0) {6};
    \node (7) at (1,0) {7};
    \node (8) at (2,0) {8};
\end{scope}

\definecolor{lightlightgray}{rgb}{0.85,0.85,0.85}

\ifedges
  \newcommand{\edgecol}{black}
\else
  \newcommand{\edgecol}{lightlightgray}
\fi

\begin{scope}[every node/.style={fill=white,circle},
              every edge/.style={\edgecol}]
    \draw[-,\edgecol] (0) -- (1);
    \draw[-,\edgecol] (0) -- (4);
    \draw[-,\edgecol] (0) -- (3);
    \draw[-,\edgecol] (1) -- (3);
    \draw[-,\edgecol] (1) -- (4);
    \draw[-,\edgecol] (1) -- (5);
    \draw[-,\edgecol] (1) -- (2);
    \draw[-,\edgecol] (2) -- (4);
    \draw[-,\edgecol] (2) -- (5);
    \draw[-,\edgecol] (3) -- (4);
    \draw[-,\edgecol] (3) -- (7);
    \draw[-,\edgecol] (3) -- (6);
    \draw[-,\edgecol] (4) -- (6);
    \draw[-,\edgecol] (4) -- (7);
    \draw[-,\edgecol] (4) -- (8);
    \draw[-,\edgecol] (4) -- (5);
    \draw[-,\edgecol] (5) -- (7);
    \draw[-,\edgecol] (5) -- (8);
    \draw[-,\edgecol] (6) -- (7);
    \draw[-,\edgecol] (7) -- (8);

    \draw[-,\edgecol]  (0) to [out=45,in=135] (2);
    \draw[-,\edgecol]  (3) to [out=30,in=150] (5);
    \draw[-,\edgecol]  (6) to [out=315,in=225] (8);
    \draw[-,\edgecol]  (0) to [out=225,in=135] (6);
    \draw[-,\edgecol]  (2) to [out=315,in=45] (8);
    \draw[-,\edgecol]  (1) to [out=300,in=60] (7);
\end{scope}
    
    \draw[orange!80, very thick, rounded corners=3mm, opacity=\op] (-0.4,2.4) -- (1.4,2.4) -- (1.4,0.6) -- (-0.4,0.6) -- cycle;

    \draw[<-,orange!80,thick] (0,0.6) -- (6);
    \draw[<-,orange!80,thick] (1,0.6) -- (7);
    \draw[<-,orange!80,thick] (1.3,0.7) -- (8);
    \draw[<-,orange!80,thick] (1.4,1) -- (5);
    \draw[<-,orange!80,thick] (1.4,2) -- (2);
    


    \end{tikzpicture}\hspace*{0.105cm}
    }%
    \hspace{\viggap}%
    \caption{VIG factorizations and derived conditional distributions, on the \textit{REBGrid} problem ($\ell=9$). Shown are the original MCond partitioning technique (starting from randomly selected vertex 8), in (a) and (b), and the proposed clique seeding technique, in (c) and (d).}
    \label{fig:reb-grid-vig}
\end{figure}

    
    




    

\section{Learning the Problem Structure}
\label{sec:learning}

In some real-world problems, the VIG of a problem may not be known.
Even if it is, not all direct dependencies encoded by a generic sub-function definition $\SUB$ may actually be of sufficient importance during optimization, making it desirable to learn the VIG during optimization.
In this section, we describe how variable dependencies can be learned within RV-GOMEA, and can then be used to learn a VIG and an FOS linkage model.
Finally, a new linkage model is proposed to better reflect overlapping dependency cliques.


\subsection{Learning Variable Dependencies}
\label{sec:learning:dependencies}

To construct a VIG during optimization, information about direct dependencies between variables is needed.
Previous work has shown that this information can be collected in a Dependency Strength Matrix (DSM) of size $\ell \times \ell$, using fitness dependency strength tests between variables~\cite{Olieman2021}.
A DSM is constructed by taking each possible pair of variables $i,j \in \IND$ and testing the strength of their dependency.
A dependency is present when $| \Delta_i - \Delta_{i,j} | \geq \epsilon$ for some small $\epsilon$, with $\Delta_i$ and $\Delta_{i,j}$ being defined as:
\begin{align}
    \Delta_i &= (f(\bm{x}) | \bm{x}_i = a_i, \bm{x}_j = a_j) \\
             &- (f(\bm{x}) | \bm{x}_i = a_i + b_i, \bm{x}_j = a_j), \nonumber \\
    \Delta_{i,j} &= (f(\bm{x}) | \bm{x}_i = a_i, \bm{x}_j = a_j + b_j) \\
             &- (f(\bm{x}) | \bm{x}_i = a_i + b_i, \bm{x}_j = a_j + b_j), \nonumber
\end{align}
where $a_i$ and $b_i$ are chosen randomly such that $x_i$ and $x_j$ stay within function bounds.
The dependency strength is then defined as:
\begin{equation}
    d_{i,j} = \begin{cases}
        1 - \frac{\Delta_{i,j}}{\Delta_i}, & \text{if } \Delta_i \geq \Delta_{i,j}\\
        1 - \frac{\Delta_i}{\Delta_{i,j}}, & \text{otherwise.}
    \end{cases}
\end{equation}
Although the strengths in a DSM are not subject to an absolute order, relative differences between $d_{i,j}$ and $d_{i,k}$ can be compared.
Dependencies with a strength below a cut-off value of $\eta=10^{-6}$ are considered negligible and assigned a strength of 0.

Constructing a complete DSM requires $(\ell * (\ell-1))/2$ fitness dependency tests.
In the absence of partial evaluations, this requires a quadratic overhead of function evaluations to perform, as each test consists of four evaluations.
Partial evaluations can make it possible to decrease this overhead to a linear order~\cite{Olieman2021}, depending on the size of sub-functions.
These fitness dependency test evaluations could all be performed at the start, but as many real-world GBO problems can be assumed to have sparse DSMs, it is desirable to spread this computational load across time, thereby biasing this method slightly towards problems which are not fully dependent.
Moreover, dependency strengths may change during optimization, further motivating an incremental dependency learning method.
Such a method has been proven effective at discovering DSMs within RV-GOMEA during optimization~\cite{Olieman2021}.
Within this method, a scheduling mechanism is used to limit the number of simultaneous checks and to pause temporarily if no dependencies are found.
An example of a DSM constructed from fitness dependency tests, using this method in RV-GOMEA, can be seen in Figure~\ref{fig:reb-grid-example:dsm}.

\begin{figure}
    \centering
    \includegraphics[width=\linewidth]{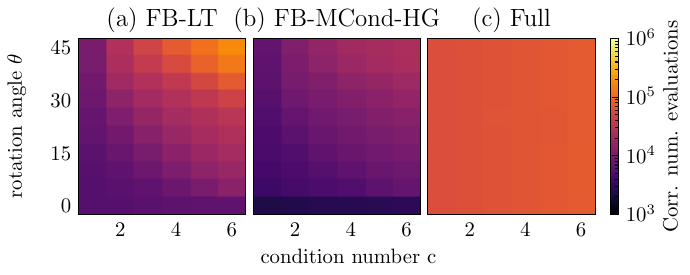}
    \caption{Required corrected number of evaluations for different linkage models on $f^{REB}(\bm{x}, c, \theta, k=2, s=1)$ with $\ell=20$, as determined by bisection. The median of 5 repeats is shown.}
    \label{fig:lt-vs-mcond}
\end{figure}

In this work, we remove one of the pause conditions of the scheduling mechanism.
In the previously proposed method, if no dependencies are found in one test iteration (consisting of $\ell$ dependency tests), an exponentially sized dependency searching pause is added to the schedule.
Preliminary experiments have indicated that this can lead to premature pauses on problems with very sparse DSMs and weak dependencies, as an unfortunate random draw of tested variable pairs can result in zero dependencies being found and the discovery being paused.
Removing this condition prevents these premature pauses while not significantly increasing the learning overhead for problems without dependencies.
The other pause condition, which tests overall dependency discovery performance across a longer interval, is kept in place.

\subsection{Learning a VIG from a DSM}

We assume that each non-zero dependency strength in the learned DSM represents a direct dependency.
The direct dependencies derived in this way encode a VIG, which can then be used to build conditional models such as the MCond-HG model described earlier.
In this work, we rebuild the VIG in every generation in which the DSM has changed, meaning that the VIG improves in quality along with the DSM, as the optimization progresses.

A preliminary experiment is performed to explore the effectivity of this method on problems with variable dependencies of increasing strength.
We consider the Rotated Ellipsoid Block (REB) problem function, which is defined in Equation~\ref{eq:reb}.
This problem is defined using blocks of the well-known rotated ellipsoid problem, in which every variable is a dependent on every other variable. 
These sub-functions then may overlap, depending on the parameters of the problem.
As the rotation angles and condition numbers of the ellipsoidal sub-functions increase, the strength of dependencies between variables, and thereby the problem difficulty, increases.
In Figure~\ref{fig:lt-vs-mcond}, we show the required number of evaluations to solve this problem to a value of $10^{-10}$ with $\ell=20$, using a bisection process described in Section~\ref{sec:experiments:setup}.
The figure compares a fitness-based linkage tree FOS model~\cite{Olieman2021}, a fitness-based conditional MCond-HG model (this work), and a full FOS model as baseline.
Both the linkage tree and conditional model can successfully be constructed from an incrementally constructed DSM, but it is clear that the conditional model scales far better than the linkage tree model.
The full model requires a stable number of required evaluations, regardless of the dependency strengths.
However, this model is known not to scale well to higher dimensions, as well as not to exploit the GBO nature of the problem at all.

\subsection{Detecting Overlapping Structures}

The MCond-HG linkage model is constructed using a breadth-first traversal partitioning strategy.
If conditional dependencies are disregarded, this process produces a (non-overlapping) marginal product FOS model, even if there are overlapping cliques of jointly dependent variables.
The disjoint nature of this model is needed for a valid forward sampling process of the factorized conditional distribution, but leads to a number of overlapping cliques not being directly modelled.
In GBO problems with sparse DSMs, it may be beneficial to model these overlapping cliques, independently, to treat all clique sub-structures, equally.
Of note: this is the case because the GOM variation operator is used that tests for an improvement after each sampling step, which can therefore break clique sub-structures not modelled as such.
At the GG optimization level, the MCond and UCond factorizations are probabilistically equivalent.

We propose an amended version of the MCond-HG model by replacing the FG step with a potentially overlapping clique FOS model.
This model is built using a technique we call \textit{clique seeding}.
This technique starts with a set of breadth-first clique searches, with independent traversals started from each variable. 
For each of these searches, the maximal candidate clique $\CLI_i'$ surrounding the start-variable is kept.
From all candidate cliques, any cliques $\CLI_i'$ are removed for which $\CLI_i' = \CLI_j'$ holds for another candidate clique $\CLI_j'$.
The final set of cliques is used as the conditional FOS model for the FG step, with the parent variables of the clique being all variables outside the clique that any variable inside the clique has a dependency with.
Each maximal clique with its dependencies effectively forms a local conditional probability distribution.
Although this can result in more dependencies being modelled than strictly necessary (there may be variables on which only some, but not all clique members depend on), this technique ensures that all maximal cliques are modelled.
This effectively increases the probability that all variables pertaining to a single sub-function are jointly sampled, which may be very important to achieve improved efficiency in case of sub-functions with strongly dependent variables.

In Figure~\ref{fig:reb-grid-vig:new}, we show an example of the maximal cliques found by the new clique seeding technique, and compare this to the disjoint cliques found by the existing partitioning technique (see Figure~\ref{fig:reb-grid-vig:old}).
The found maximal cliques are:
\begin{equation*}
    \begin{gathered}
        \bm{X}_{\{0, 1, 3, 4\}}, 
        \bm{X}_{\{1, 2, 4, 5\}}, 
        \bm{X}_{\{3, 4, 6, 7\}}, 
        \bm{X}_{\{4, 5, 7, 8\}}, \\
        \bm{X}_{\{ 0, 1, 2, 4 \}}, 
        \bm{X}_{\{ 2, 4, 5, 8 \}}, 
        \bm{X}_{\{ 4, 6, 7, 8 \}}, 
        \bm{X}_{\{ 0, 3, 4, 6 \}}, 
        \bm{X}_{\{ 1, 3, 4, 5, 7 \}}.
    \end{gathered}
\end{equation*}
The derived local conditional distribution of one maximal clique is shown in Figure~\ref{fig:reb-grid-vig:new:network}.
In RV-GOMEA, we use the new clique seeding method alongside the existing partitioning technique.
The GG step, which samples from the factorized conditional distribution, still uses the original MCond factorization, as this is guaranteed to be consistent across all variables.
This global sampling step can serve to restore lost linkage where overlaps exist.

\begin{table}
    \small
    \centering
    \setlength{\tabcolsep}{4pt}
    \renewcommand{\arraystretch}{1.25}
    \begin{tabularx}{\linewidth}{lX}
        \toprule
        \textbf{Linkage model} & \textbf{Description} \\
        \midrule
        Univariate & Each variable modeled independently \\
        Static-UCond-HG & Pre-defined conditional univariate FOS with a full forward sampling step~\cite{Bouter2020b} \\
        Static-MCond-HG & Pre-defined conditional marginal product FOS with a full forward sampling step~\cite{Bouter2020b} \\
        Static-MCond-HG-CS & Static-MCond-HG with a clique-seeded FOS \textbf{($\ast$)} \\
        FB-LT & Fitness-based bounded, pruned linkage tree~\cite{Olieman2021}  \\
        FB-UCond-HG & Fitness-based conditional univariate FOS with a full forward sampling step \textbf{($\ast$)} \\
        FB-MCond-HG & Fitness-based conditional marginal product FOS with a full forward sampling step \textbf{($\ast$)} \\
        FB-MCond-HG-CS & FB-MCond-HG with a clique-seeded FOS \textbf{($\ast$)} \\
        \bottomrule
    \end{tabularx}
    \caption{Overview of compared linkage models. Models marked with an asterisk {\small \textbf{($\ast$)}} are proposed in this work.}
    \label{tab:linkage-models}
\end{table}

\vspace{-0.1cm}

\section{Experiments}
\label{sec:experiments}

In this section, we evaluate and compare the performance of RV-GOMEA with the proposed learned conditional linkage models to RV-GOMEA with previously proposed models.
Table~\ref{tab:linkage-models} lists all linkage models, which include generic models, static (pre-defined) models, and fitness-based (learned) models. 
We use the UCond-HG and MCond-HG variants as representatives of static conditional linkage models~\cite{Bouter2020b} and the pruned linkage tree as a representative of fitness-based non-conditional models~\cite{Olieman2021}.

We also compare the performance of RV-GOMEA to VkD-CMA~\cite{Akimoto2016,Akimoto2016a}, which is a state-of-the-art optimization method for continuous optimization problems.
VkD-CMA is also suited for problems with a sparse DSM, but cannot leverage partial problem evaluations due to its design.

\vspace{-0.1cm}

\subsection{Benchmark Problems}
\label{sec:experiments:problems}

We conduct this comparison on a set of benchmark problems that allow for partial evaluations.
Both separable and non-separable problems are included, to verify if the fitness dependency detection mechanism included in several linkage models is robust to both the absence and presence of separable sub-structures.

As a decomposable baseline problem, we consider the well-known \textit{Sphere} function.
This problem is defined as:
\begin{equation}
    f^{\textit{Sphere}}(\bm{x}) = \sum_{i=0}^{\ell-1} \left[ \bm{x}_i^2 \right] .
\end{equation}

The second problem we consider is the \textit{Rosenbrock} function:  
\begin{equation}
    f^{\textit{Rosenbrock}}(\bm{x}) = \sum_{i=0}^{\ell-2} \left[ 100(\bm{x}_{i+1} - \bm{x}_i^2)^2 + (1 - \bm{x}_i)^2 \right] .
\end{equation}

The remaining problems are all derived from rotated ellipsoid block (REB) functions, defined as:
\begin{align}
    \label{eq:ellipsoid}
    f^{\textit{Ellipsoid}}(\bm{x}, c) &= \sum_{i=0}^{|\bm{x}| - 1} \left[ 10^{c*i/(|\bm{x}| - 1)} \bm{x}^2_i \right] , \\
    \label{eq:reb}
    f^{\textit{REB}}(\bm{x}, c, \theta, k, s) &= \sum_{i=0}^{\lceil \frac{|\bm{x}|-k}{s} \rceil - 1} f^{\textit{Ellipsoid}}(R^{\theta} (\bm{x}_{[is : is + k - 1]}), c) ,
\end{align}
where $c$ defines the condition number of the REBs, $\theta$ the rotation angle with which they are rotated, $k$ their size, and $s$ the stride $1 \leq s$ with which they are spaced.
The rotation function $R^{\theta}(\bm{x})$ rotates $\bm{x}$ by $\theta$ degrees around the origin, counter-clockwise.

REB functions provide us with fine-grained control over the degree of overlap and difficulty of optimization problems.
Large condition numbers $c$ and rotation angles $\theta$ induce strong dependencies between variables grouped together by blocks.

We derive the following problem functions from $f^{REB}$:
\begin{align}
    f^{\textit{REB2Weak}}(\bm{x}) &= f^{\textit{REB}}(\bm{x}, c=1, \theta=5, k=2, s=1) , \\
    f^{\textit{REB2Strong}}(\bm{x}) &= f^{\textit{REB}}(\bm{x}, c=6, \theta=45, k=2, s=1) , \\
    f^{\textit{REB5NoOverlap}}(\bm{x}) &= f^{\textit{REB}}(\bm{x}, c=6, \theta=45, k=5, s=5) , \\
    f^{\textit{REB5SmallOverlap}}(\bm{x}) &= f^{\textit{REB}}(\bm{x}, c=6, \theta=45, k=5, s=4) , \\
    f^{\textit{REB5LargeOverlap}}(\bm{x}) &= f^{\textit{REB}}(\bm{x}, c=6, \theta=45, k=5, s=1).
\end{align}

\begin{figure*}
    \setlength{\belowcaptionskip}{-3pt}
    \centering
    \includegraphics[width=\textwidth]{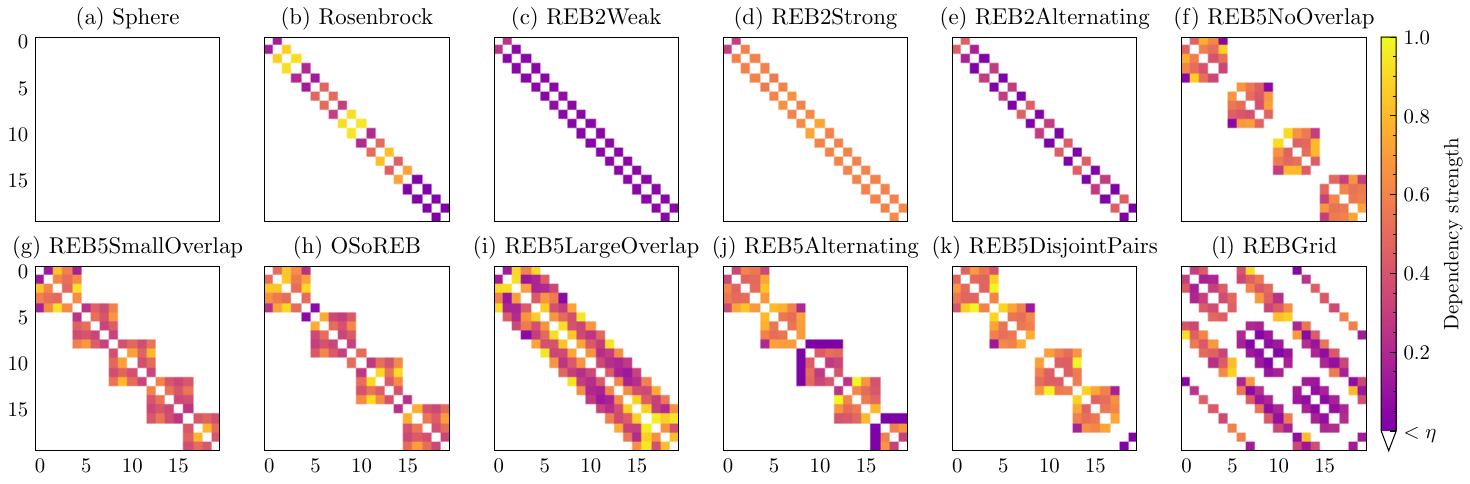}
    \caption{Heatmaps depicting a part ($\IND_{[0:19]} \times \IND_{[0:19]}$) of the computed DSM for each tested problem at a dimensionality $\ell > 30$. Variable pairs which are deemed independent are depicted in white.}
    \label{fig:dsm-grid}
\end{figure*}

The \textit{OSoREB} problem, which previous work has optimized using the FB-LT linkage model~\cite{Olieman2021}, is also included:
\begin{equation}
\begin{split}
    f^{\textit{OSoREB}}(\bm{x}) &= f^{\textit{REB}}(\bm{x}, c=6, \theta=45, k=5, s=4) \\
                                 &+ f^{\textit{REB}}(\bm{x}_{[4:|\bm{x}| - 1]}, c=6, \theta=45, k=2, s=5).
\end{split}
\end{equation}

To test heterogeneous block strengths, we also define:
\begin{align}
    f^{\textit{REB2Alternating}}(\bm{x}) &= f^{\textit{REB}}(\bm{x}, c_i, \theta_i, k=2, s=1) , \\
    f^{\textit{REB5Alternating}}(\bm{x}) &= f^{\textit{REB}}(\bm{x}, c_i, \theta_i, k=5, s=4) , \\ %
    \text{with } c_i=1, \theta_i=5 \text{ if } i &\text{ is even, otherwise } c_i=6, \theta_i=45 , \nonumber \\%
    f^{\textit{REB5DisjointPairs}}(\bm{x}) &= f^{\textit{REB}}(\bm{x}, c=6, \theta=45, k=5, s_i) , \\
    \text{with } s_i=4 \text{ if } i &\text{ is even, otherwise } s_i=5 . \nonumber
\end{align}

Finally, we consider a more tightly connected REB function, where variables are arranged as vertices $v \in \bm{V}$ in a square grid of size $\sqrt{\ell} \times \sqrt{\ell}$.
Horizontally and vertically neighboring vertices ($a \in N(b)$) are connected by an edge.
This problem is defined as:
\begin{equation}
    f^{\textit{REBGrid}}(\bm{x}) = \sum_{v \in \bm{V}} f^{\textit{Ellipsoid}}(R^{45} (\bm{x}_{\{v\} \cup N(v)}), 6) .
\end{equation}

\subsection{Experimental Setup}
\label{sec:experiments:setup}

On all problems, the optimal solution is the origin. 
All variables of individuals in each population are initialized uniformly randomly in the interval $[-115,-110]$.
We use an evaluation budget of $10^7$ full evaluations and a computation time budget of 3 hours.
Whenever either of these budgets is exhausted without finding an individual with an objective value smaller than or equal to the value-to-reach ($10^{-10}$), the run is considered failed.
If premature convergence occurs within either of these budgets, the population is restarted with the same size. 
In the interest of reproducibility, all non-deterministic factors are seeded.
Source code of the modified RV-GOMEA version, implemented in C++, along with code to reproduce the experiments in this paper, is provided on GitHub\footnote{\url{https://github.com/gandreadis/conditional-rv-gomea}}.

For each experimental configuration (consisting of a problem, a linkage model, and a dimensionality) we conduct a bisection search for the population size that yields the lowest expected number of evaluations.
During this search, each tested population size is repeated 30 times.
As some of these repeats may fail, we use as a metric the mean required number of evaluations of all successful repeats, divided by the fraction of repeats that are successful.
The bisection search is limited to population sizes between 8 and 2048, and starts at a population size of $17 + 3*\ell^{1.5}$, following a previously established guideline for the full FOS linkage model~\cite{Bosman2013}.
The size is then decreased exponentially while the required number of evaluations also monotonically decreases, to form a bracket for bisection.
We repeat this bisection 5 times and report the median outcome.

We perform scalability experiments by testing four dimensionalities: 10, 20, 40, and 80.
On problems with additional dimensionality constraints (e.g., \textit{REBGrid}), we choose a close compatible dimensionality.
As bisection on large-dimensional problems can require large amounts of computation time, the population sizes for two larger dimensionalities (160 and 320) are estimated after all bisections have completed.
We use linear extrapolation on a log-log scale to estimate the (non-decreasing) slope of required population sizes.
Using this slope, we estimate population sizes for the two larger dimensionalities and execute 5 sets of 30 repeats for each.

\begin{figure*}
    \setlength{\belowcaptionskip}{-3pt}
    \centering
    \includegraphics[width=0.95\textwidth]{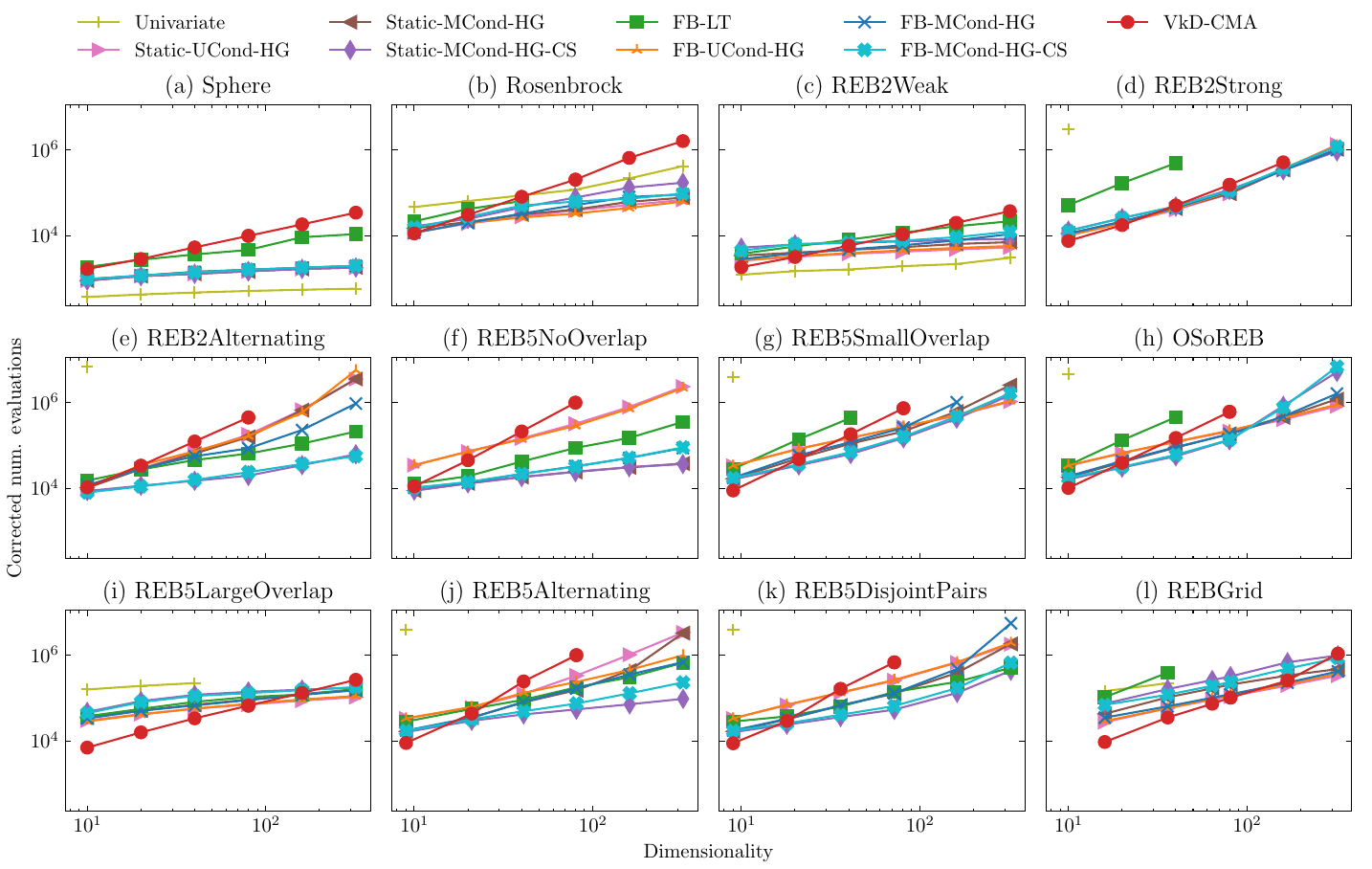}
    \vspace{-0.2cm}
    \caption{Results of scalability experiments for the tested benchmark problems and optimization approaches at different dimensionalities, as determined by bisection. The mean over 5 repeated bisections is shown.}
    \label{fig:bisection}
\end{figure*}


\section{Results}
\label{sec:results}

We visualize the DSMs of all problems in Figure~\ref{fig:dsm-grid}, as computed during optimization.
As fitness dependency accuracy can deteriorate near convergence, the shown DSMs are automatically selected from a generation before convergence.
All shown dependencies are consistently detected across 30 runs, albeit with different relative strengths.
While we only show DSMs of one run here, the DSMs of another run and the average over 30 runs are provided in Appendix~\ref{sec:dsms}.
We conclude from visual inspection that the problem structure is successfully learned during optimization.

In Figure~\ref{fig:bisection}, we show the scalability of the corrected mean number of evaluations needed by VkD-CMA and the compared linkage models in RV-GOMEA.
The corresponding population sizes, and the outcomes of statistical significance tests, are included Appendices~\ref{sec:population-sizes} and~\ref{sec:statistical-significance}.
We find that the fitness-based conditional linkage models scale similarly to their static counterparts, with only little overhead and with resilience against the absence of dependencies.
As expected, on separable problems such as \emph{REB5NoOverlap}, the MCond-HG models and their clique-seeding variants perform equally well, as their models are equivalent.
On multiple problems with overlapping structures and heterogeneous dependency strengths, such as \emph{REB5Alternating} and \emph{REB2Alternating}, the proposed static clique seeding model scales better than existing static models ($p<0.01$).
On problems with less distinct structures, such as \emph{REBGrid} and \emph{REB5LargeOverlap}, the different conditional models scale similarly.
Across the large majority of problems, conditional linkage models outperform VkD-CMA at scale.


\section{Discussion}
\label{sec:discussion}

The results of this scalability analysis clearly show it is possible to learn the structure of a GBO problem while optimizing it, with little overhead.
Moreover, RV-GOMEA with the newly introduced clique seeding technique can scale better on several problems than RV-GOMEA with the previously introduced conditional linkage models.
However, there does not appear to be one clearly superior linkage model for all different dependency structures.
This raises the question of how to choose the appropriate linkage model for a particular problem, especially if its structure is not known a priori.

One characteristic which may inform the choice of linkage model could be the relative strengths of connected dependency structures.
Clear differences can be observed on the three \emph{REB2} problems, which differ only in their dependency strengths:
On \emph{REB2Weak} and \emph{REB2Strong}, both featuring homogeneous dependency strengths, scalability is similar for most linkage models, while on \emph{REB2Alternating}, where strong and weak blocks are interleaved, clique seeding models outperform the other models.
Similar observations can be made on \emph{REB5Alternating} and \emph{REB5SmallOverlap}.
These differences could be explained by the differing FOS models: whereas an MCond factorization may capture only some strong blocks directly as FOS elements due to unfortunate partitioning, the clique seeding technique models all blocks separately.
In future work, the distinctness of dependency structures, i.e., the heterogeneity of their strengths, could therefore be an indicator for automatic linkage model selection.
Furthermore, dependency strength could be modelled in a probabilistic fashion, e.g., with a dynamic cut-off dependency strength value.

The experiments in this work are limited to artificial benchmark problems.
Although these allow for fine-grained control over dependency structures, many real-world problems feature less homogeneous structures.
One example is the medical deformable image registration problem~\cite{Andreadis2023}, which RV-GOMEA has shown to be capable of efficiently optimizing.
Currently, the FOS model used in that work is a generic, non-conditional model based on locality as a heuristic for dependency strength.
The fitness dependency learning and conditional modelling techniques presented here therefore would likely ensure a more tailored and efficient linkage model.

Finally, in this work we use an existing technique for fitness dependency learning, which is based on pairwise variable checks.
A promising direction of future research could be the incorporation of more efficient, hierarchical fitness-dependency learning techniques~\cite{Komarnicki2023}, potentially further reducing the learning overhead.


\vspace{-0.15cm}

\section{Conclusions}
\label{sec:conclusions}

In this paper, we set out to extend RV-GOMEA's excellent performance on GBO problems with strong overlapping dependencies, to the case where the VIG is not known beforehand.
This was accomplished by applying fitness-based linkage learning techniques to construct the VIG during optimization, and deriving conditional linkage models from this VIG.
Moreover, we proposed a new linkage modeling technique for overlapping structures.
The efficiency of these learned conditional linkage models was compared to static linkage models and learned non-conditional linkage models in RV-GOMEA, and to VkD-CMA, a state-of-the-art algorithm for continuous optimization.
This comparison was conducted on benchmark GBO problems with varying dependency structures.

Results of this comparison showed that the learning of a conditional linkage model can be done accurately and with negligible overhead.
On the majority of problems, learned conditional linkage models scaled equally well as, or better than VkD-CMA.
Furthermore, the newly introduced clique seeding technique demonstrated superior performance to existing partitioning techniques on several problems.
The proposed linkage models can be applied to real-world optimization problems where the VIG is not known beforehand.
This can also give valuable insights into problem structure, facilitating knowledge transfer between problem instances.

\vspace{-0.2cm}
\begin{acks}
This research is part of the Open Technology Programme with project number 15586, financed by the Dutch Research Council (NWO), Elekta, and Xomnia. 
Further, this work is co-funded by the public-private partnership allowance for top consortia for knowledge and innovation (TKIs) from the Ministry of Economic Affairs.
\end{acks}

\appendix

\section{Analysis of Dependency Detection Performance}
\label{sec:dsms}

\begin{figure*}
    \centering
    
    \begin{subfigure}[b]{\textwidth}
       \includegraphics[width=\linewidth]{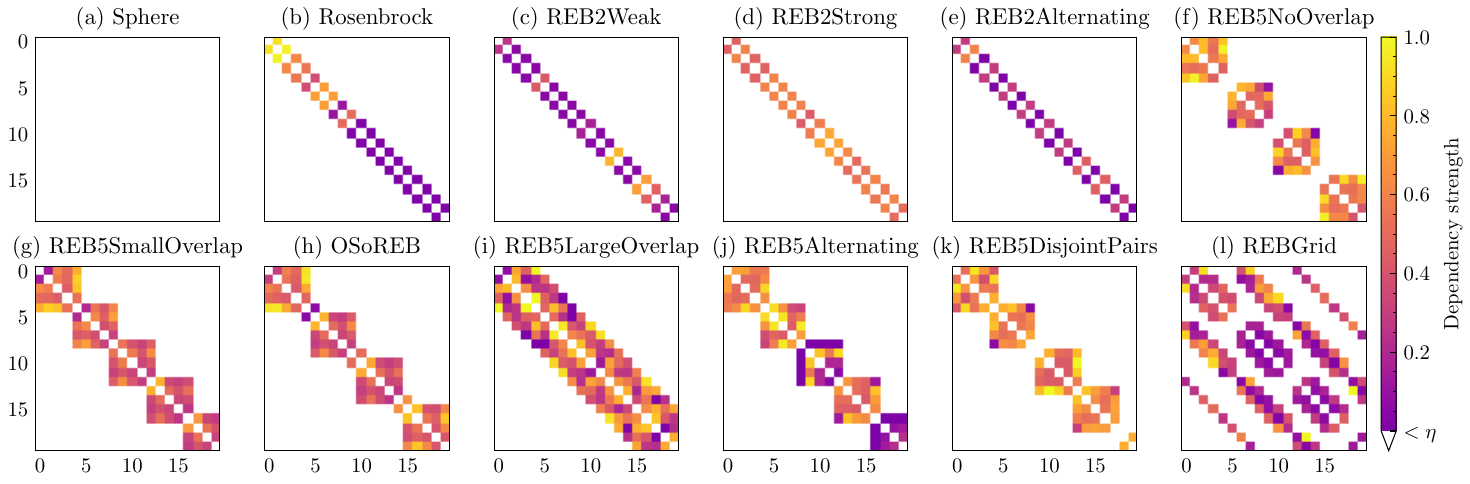}
       \caption{DSMs obtained during a second run.}
       \label{fig:dsm:repeat} 
    \end{subfigure}
    
    \vspace{0.1cm}
    
    \begin{subfigure}[b]{\textwidth}
       \includegraphics[width=\linewidth]{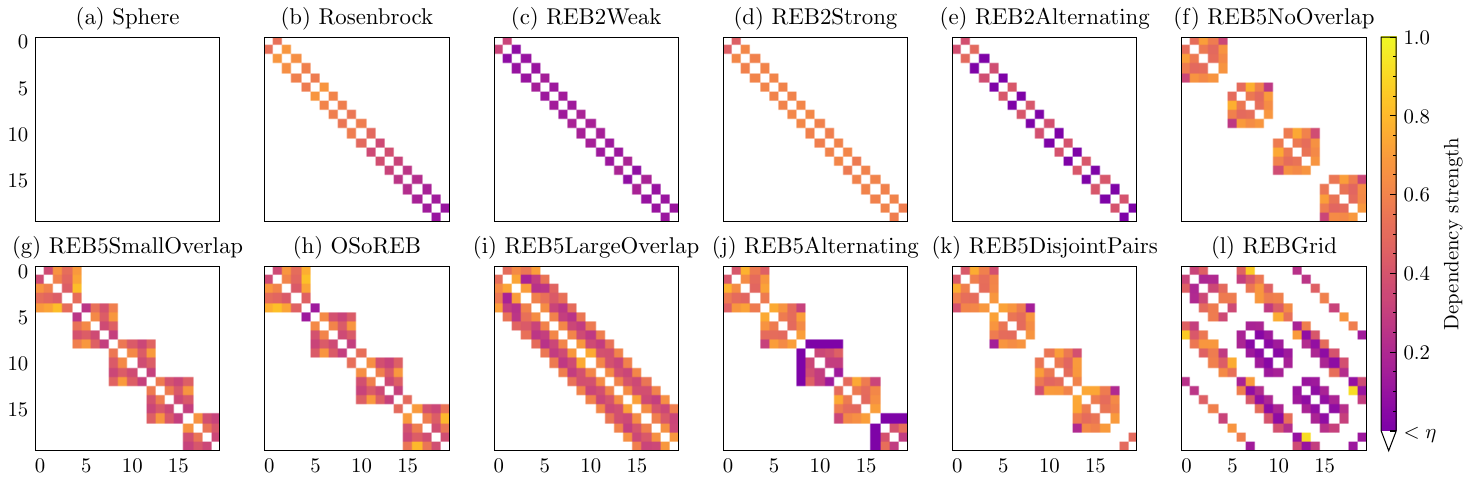}
       \caption{DSMs averaged across 30 runs.}
       \label{fig:dsm:average}
    \end{subfigure}
    \vspace{-0.1cm}

    \caption{Additional heatmaps of computed DSMs, for a second run and an average of 30 runs.}
    \label{fig:dsm:extended}
\end{figure*}

To test the reliability of the incremental DSM construction mechanism described in Section~\ref{sec:learning:dependencies}, additional runs were performed.
Next to the DSMs of one run per problem, shown in Figure~\ref{fig:dsm-grid}, the DSMs of a second run per problem are shown in Figure~\ref{fig:dsm:repeat}.
As expected, the same dependencies are found, but with different relative strengths, since these depend on the location and spread of the population.
In addition, we show the average DSMs over 30 runs per problem in Figure~\ref{fig:dsm:average}.
This follows earlier work which uses a similar aggregation technique for visualization purposes~\cite{Olieman2021}.
We observe that the average relative strengths are generally similar to the relative strengths in individual runs, but that the average DSMs tend to show less local heterogeneity.

\section{Population Sizing Results}
\label{sec:population-sizes}

\begin{figure*}
    \centering
    \includegraphics[width=\textwidth]{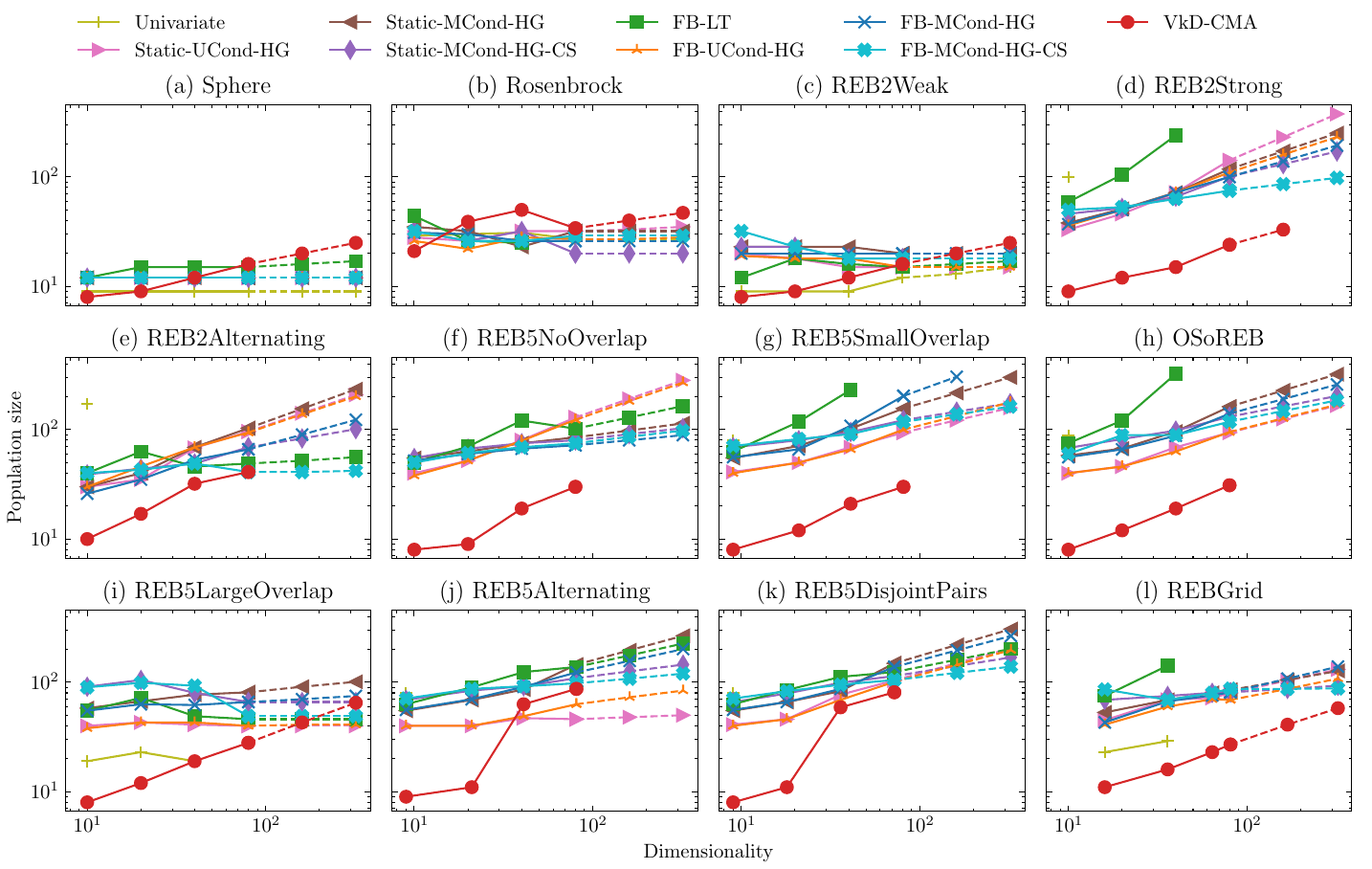}
    \caption{Minimum required population sizes, as determined by scalability experiments for the tested benchmark problems and optimization approaches at different dimensionalities. The mean over 5 repeated bisections is shown. The population sizes of the two largest dimensions of each problem are determined by extrapolation.}
    \label{fig:bisection:population}
\end{figure*}

Figure~\ref{fig:bisection:population} shows the population sizes derived from the bisection and extrapolation process described in Section~\ref{sec:experiments:setup}.
In general, similar general trends can be observed as in the scalability of required evaluations (see Figure~\ref{fig:bisection}), although with notable differences.
VkD-CMA generally requires smaller population sizes than the RV-GOMEA variants, which contrasts with its scalability in terms of required evaluations.
Furthermore, we observe that many population size scalability curves are non-monotonic, while the same curves for the required number of evaluations are typically monotonic.
This emphasizes the need to use the required number of evaluations as the main minimization objective in bisection-based population sizing schemes.

\section{Statistical Significance of Results}
\label{sec:statistical-significance}

We perform statistical significance tests to test if the evaluation scalability results in Figure~\ref{fig:bisection} are significant.
For each problem, at its largest dimensionality, we take the required number of evaluations of each model, i.e., a linkage model of RV-GOMEA or VkD-CMA.
We choose the model with the best median scalability and compare the five repeats of that model against the repeats of the other models, in a series of Mann-Whitney U tests.
To ensure fair tests, we split the models into two categories, based on whether they received VIG knowledge, \textit{a priori}.
Table~\ref{tab:statistical-tests} lists the results of these tests, with multiple-test correction applied.
We conclude that the majority of visually observed differences in scalability behavior is also statistically significant.

\begin{table}
    \small
    \centering
    \setlength{\tabcolsep}{4pt}
    \begin{tabularx}{\linewidth}{clrlr}
        \toprule
        Problem  & \multicolumn{2}{c}{Best with VIG knowledge} & \multicolumn{2}{c}{Best without VIG knowledge} \\
        & Model & $p$ & Model & $p$ \\
        \midrule
(a) & Static-UCond-HG & 1.000 & Univariate & {\bftab 0.008} \\
(b) & Static-UCond-HG & 0.310 & FB-UCond-HG & 0.036 \\
(c) & Static-UCond-HG & {\bftab 0.008} & Univariate & {\bftab 0.008} \\
(d) & Static-MCond-HG-CS & 0.032 & FB-MCond-HG & 0.056 \\
(e) & Static-MCond-HG-CS & {\bftab 0.008} & FB-MCond-HG-CS & 0.016 \\
(f) & Static-MCond-HG-CS & {\bftab 0.016} & FB-MCond-HG & {\bftab 0.008} \\
(g) & Static-UCond-HG & {\bftab 0.008} & FB-UCond-HG & {\bftab 0.008} \\
(h) & Static-UCond-HG & {\bftab 0.008} & FB-UCond-HG & {\bftab 0.008} \\
(i) & Static-UCond-HG & {\bftab 0.008} & FB-UCond-HG & {\bftab 0.008} \\
(j) & Static-MCond-HG-CS & {\bftab 0.008} & FB-MCond-HG-CS & {\bftab 0.008} \\
(k) & Static-MCond-HG-CS & {\bftab 0.008} & FB-LT & {\bftab 0.008} \\
(l) & Static-UCond-HG & {\bftab 0.008} & FB-UCond-HG & 0.190 \\
        \bottomrule
    \end{tabularx}
    \caption{Statistical tests for the largest tested dimensionality. The model with the lowest median corrected number of evaluations is tested against the other models. Reported is the highest $p$-value of all tests, at a Bonferroni-adjusted $p$-level of $0.05/n$, with $n$ being the number of tests.}
    \label{tab:statistical-tests}
\end{table}

\newpage~
\newpage~
\newpage~
\newpage~
\newpage~
\newpage

\bibliographystyle{ACM-Reference-Format}
\bibliography{references}


\begin{thebibliography}{29}


\ifx \showCODEN    \undefined \def \showCODEN     #1{\unskip}     \fi
\ifx \showDOI      \undefined \def \showDOI       #1{#1}\fi
\ifx \showISBNx    \undefined \def \showISBNx     #1{\unskip}     \fi
\ifx \showISBNxiii \undefined \def \showISBNxiii  #1{\unskip}     \fi
\ifx \showISSN     \undefined \def \showISSN      #1{\unskip}     \fi
\ifx \showLCCN     \undefined \def \showLCCN      #1{\unskip}     \fi
\ifx \shownote     \undefined \def \shownote      #1{#1}          \fi
\ifx \showarticletitle \undefined \def \showarticletitle #1{#1}   \fi
\ifx \showURL      \undefined \def \showURL       {\relax}        \fi
\providecommand\bibfield[2]{#2}
\providecommand\bibinfo[2]{#2}
\providecommand\natexlab[1]{#1}
\providecommand\showeprint[2][]{arXiv:#2}

\bibitem[\protect\citeauthoryear{Ahn, Ramakrishna, and Goldberg}{Ahn et~al\mbox{.}}{2004}]%
        {Ahn2004}
\bibfield{author}{\bibinfo{person}{C.~W. Ahn}, \bibinfo{person}{R.~S. Ramakrishna}, {and} \bibinfo{person}{D.~E. Goldberg}.} \bibinfo{year}{2004}\natexlab{}.
\newblock \showarticletitle{{Real-coded Bayesian optimization algorithm: Bringing the strength of BOA into the continuous world}}. In \bibinfo{booktitle}{\emph{Proceedings of the 2004 Genetic and Evolutionary Computation Conference}}, Vol.~\bibinfo{volume}{3102}. \bibinfo{pages}{840--851}.
\newblock


\bibitem[\protect\citeauthoryear{Akimoto and Hansen}{Akimoto and Hansen}{2016a}]%
        {Akimoto2016a}
\bibfield{author}{\bibinfo{person}{Y. Akimoto} {and} \bibinfo{person}{N. Hansen}.} \bibinfo{year}{2016}\natexlab{a}.
\newblock \showarticletitle{{Online Model Selection for Restricted Covariance Matrix Adaptation}}. In \bibinfo{booktitle}{\emph{International Conference on Parallel Problem Solving from Nature}}. \bibinfo{pages}{3--13}.
\newblock


\bibitem[\protect\citeauthoryear{Akimoto and Hansen}{Akimoto and Hansen}{2016b}]%
        {Akimoto2016}
\bibfield{author}{\bibinfo{person}{Y. Akimoto} {and} \bibinfo{person}{N. Hansen}.} \bibinfo{year}{2016}\natexlab{b}.
\newblock \showarticletitle{{Projection-based restricted covariance matrix adaptation for high dimension}}. In \bibinfo{booktitle}{\emph{Proceedings of the 2016 Genetic and Evolutionary Computation Conference}}. \bibinfo{pages}{197--204}.
\newblock


\bibitem[\protect\citeauthoryear{Alden and Miikkulainen}{Alden and Miikkulainen}{2016}]%
        {Alden2016}
\bibfield{author}{\bibinfo{person}{M. Alden} {and} \bibinfo{person}{R. Miikkulainen}.} \bibinfo{year}{2016}\natexlab{}.
\newblock \showarticletitle{{MARLEDA: Effective distribution estimation through Markov random fields}}.
\newblock \bibinfo{journal}{\emph{Theoretical Computer Science}}  \bibinfo{volume}{633} (\bibinfo{year}{2016}), \bibinfo{pages}{4--18}.
\newblock


\bibitem[\protect\citeauthoryear{Andreadis, Bosman, and Alderliesten}{Andreadis et~al\mbox{.}}{2023}]%
        {Andreadis2023}
\bibfield{author}{\bibinfo{person}{G. Andreadis}, \bibinfo{person}{P.~A.~N. Bosman}, {and} \bibinfo{person}{T. Alderliesten}.} \bibinfo{year}{2023}\natexlab{}.
\newblock \showarticletitle{{MOREA: a GPU-accelerated Evolutionary Algorithm for Multi-Objective Deformable Registration of 3D Medical Images}}. In \bibinfo{booktitle}{\emph{Proceedings of the 2023 Genetic and Evolutionary Computation Conference}}. \bibinfo{pages}{1294--1302}.
\newblock


\bibitem[\protect\citeauthoryear{Bosman, Grahl, and Thierens}{Bosman et~al\mbox{.}}{2013}]%
        {Bosman2013}
\bibfield{author}{\bibinfo{person}{P.~A.N. Bosman}, \bibinfo{person}{J. Grahl}, {and} \bibinfo{person}{D. Thierens}.} \bibinfo{year}{2013}\natexlab{}.
\newblock \showarticletitle{{Benchmarking parameter-free AMaLGaM on functions with and without noise}}.
\newblock \bibinfo{journal}{\emph{Evolutionary Computation}} \bibinfo{volume}{21}, \bibinfo{number}{3} (\bibinfo{year}{2013}), \bibinfo{pages}{455--469}.
\newblock


\bibitem[\protect\citeauthoryear{Bosman}{Bosman}{2009}]%
        {Bosman2009}
\bibfield{author}{\bibinfo{person}{P.~A.~N. Bosman}.} \bibinfo{year}{2009}\natexlab{}.
\newblock \showarticletitle{{On Empirical Memory Design, Faster Selection of Bayesian Factorizations and Parameter-Free Gaussian EDAs}}. In \bibinfo{booktitle}{\emph{Proceedings of the 2009 Genetic and Evolutionary Computation Conference}}. \bibinfo{pages}{389--396}.
\newblock


\bibitem[\protect\citeauthoryear{Bouter, Alderliesten, Bel, Witteveen, and Bosman}{Bouter et~al\mbox{.}}{2018}]%
        {Bouter2018}
\bibfield{author}{\bibinfo{person}{A. Bouter}, \bibinfo{person}{T. Alderliesten}, \bibinfo{person}{A. Bel}, \bibinfo{person}{C. Witteveen}, {and} \bibinfo{person}{Peter A.~N. Bosman}.} \bibinfo{year}{2018}\natexlab{}.
\newblock \showarticletitle{{Large-scale parallelization of partial evaluations in evolutionary algorithms for real-world problems}}. In \bibinfo{booktitle}{\emph{Proceedings of the 2018 Genetic and Evolutionary Computation Conference}}. \bibinfo{pages}{1199--1206}.
\newblock


\bibitem[\protect\citeauthoryear{Bouter, Alderliesten, and Bosman}{Bouter et~al\mbox{.}}{2020a}]%
        {Bouter2020}
\bibfield{author}{\bibinfo{person}{A. Bouter}, \bibinfo{person}{T. Alderliesten}, {and} \bibinfo{person}{P.~A.~N. Bosman}.} \bibinfo{year}{2020}\natexlab{a}.
\newblock \showarticletitle{{Achieving highly scalable evolutionary real-valued optimization by exploiting partial evaluations}}.
\newblock \bibinfo{journal}{\emph{Evolutionary Computation}} \bibinfo{volume}{29}, \bibinfo{number}{1} (\bibinfo{year}{2020}), \bibinfo{pages}{129--155}.
\newblock


\bibitem[\protect\citeauthoryear{Bouter, Luong, Witteveen, Alderliesten, and Bosman}{Bouter et~al\mbox{.}}{2017a}]%
        {Bouter2017b}
\bibfield{author}{\bibinfo{person}{A. Bouter}, \bibinfo{person}{N.~H. Luong}, \bibinfo{person}{C. Witteveen}, \bibinfo{person}{T. Alderliesten}, {and} \bibinfo{person}{P.~A.~N. Bosman}.} \bibinfo{year}{2017}\natexlab{a}.
\newblock \showarticletitle{{The multi-objective real-valued gene-pool optimal mixing evolutionary algorithm}}. In \bibinfo{booktitle}{\emph{Proceedings of the 2017 Genetic and Evolutionary Computation Conference}}. \bibinfo{pages}{537--544}.
\newblock


\bibitem[\protect\citeauthoryear{Bouter, Maree, Alderliesten, and Bosman}{Bouter et~al\mbox{.}}{2020b}]%
        {Bouter2020b}
\bibfield{author}{\bibinfo{person}{A. Bouter}, \bibinfo{person}{S.~C. Maree}, \bibinfo{person}{T. Alderliesten}, {and} \bibinfo{person}{P.~A.~N. Bosman}.} \bibinfo{year}{2020}\natexlab{b}.
\newblock \showarticletitle{{Leveraging conditional linkage models in gray-box optimization with the real-valued gene-pool optimal mixing evolutionary algorithm}}. In \bibinfo{booktitle}{\emph{Proceedings of the 2020 Genetic and Evolutionary Computation Conference}}. \bibinfo{pages}{603--611}.
\newblock


\bibitem[\protect\citeauthoryear{Bouter, Witteveen, Alderliesten, and Bosman}{Bouter et~al\mbox{.}}{2017b}]%
        {Bouter2017c}
\bibfield{author}{\bibinfo{person}{A. Bouter}, \bibinfo{person}{C. Witteveen}, \bibinfo{person}{T. Alderliesten}, {and} \bibinfo{person}{P.~A.~N. Bosman}.} \bibinfo{year}{2017}\natexlab{b}.
\newblock \showarticletitle{{Exploiting linkage information in real-valued optimization with the real-valued gene-pool optimal mixing evolutionary algorithm}}. In \bibinfo{booktitle}{\emph{Proceedings of the 2017 Genetic and Evolutionary Computation Conference}}. \bibinfo{pages}{705--712}.
\newblock


\bibitem[\protect\citeauthoryear{Deb and Myburgh}{Deb and Myburgh}{2016}]%
        {Deb2016}
\bibfield{author}{\bibinfo{person}{K. Deb} {and} \bibinfo{person}{C. Myburgh}.} \bibinfo{year}{2016}\natexlab{}.
\newblock \showarticletitle{{Breaking the Billion-Variable Barrier in real-world optimization using a customized evolutionary algorithm}}. In \bibinfo{booktitle}{\emph{Proceedings of the 2016 Genetic and Evolutionary Computation Conference}}. \bibinfo{pages}{653--660}.
\newblock


\bibitem[\protect\citeauthoryear{Dickhoff, Kerkhof, Deuzeman, Creutzberg, Alderliesten, and Bosman}{Dickhoff et~al\mbox{.}}{2022}]%
        {Dickhoff2022}
\bibfield{author}{\bibinfo{person}{L.~R.~M. Dickhoff}, \bibinfo{person}{E.~M. Kerkhof}, \bibinfo{person}{H.~H. Deuzeman}, \bibinfo{person}{C.~L. Creutzberg}, \bibinfo{person}{T. Alderliesten}, {and} \bibinfo{person}{P.~A.~N. Bosman}.} \bibinfo{year}{2022}\natexlab{}.
\newblock \showarticletitle{{Adaptive objective configuration in bi-objective evolutionary optimization for cervical cancer brachytherapy treatment planning}}. In \bibinfo{booktitle}{\emph{Proceedings of the 2022 Genetic and Evolutionary Computation Conference}}. \bibinfo{pages}{1173--1181}.
\newblock


\bibitem[\protect\citeauthoryear{Echegoyen, Lozano, Santana, and Larra{\~{n}}aga}{Echegoyen et~al\mbox{.}}{2007}]%
        {Echegoyen2007}
\bibfield{author}{\bibinfo{person}{C. Echegoyen}, \bibinfo{person}{J.~A. Lozano}, \bibinfo{person}{R. Santana}, {and} \bibinfo{person}{P. Larra{\~{n}}aga}.} \bibinfo{year}{2007}\natexlab{}.
\newblock \showarticletitle{{Exact Bayesian Network Learning in Estimation of Distribution Algorithms}}. In \bibinfo{booktitle}{\emph{IEEE Congress on Evolutionary Computation}}. \bibinfo{pages}{1051--1058}.
\newblock


\bibitem[\protect\citeauthoryear{Gronau and Moran}{Gronau and Moran}{2007}]%
        {Gronau2007}
\bibfield{author}{\bibinfo{person}{I. Gronau} {and} \bibinfo{person}{S. Moran}.} \bibinfo{year}{2007}\natexlab{}.
\newblock \showarticletitle{{Optimal implementations of UPGMA and other common clustering algorithms}}.
\newblock \bibinfo{journal}{\emph{Inform. Process. Lett.}} \bibinfo{volume}{104}, \bibinfo{number}{6} (\bibinfo{year}{2007}), \bibinfo{pages}{205--210}.
\newblock


\bibitem[\protect\citeauthoryear{Hansen, M{\"{u}}ller, and Koumoutsakos}{Hansen et~al\mbox{.}}{2003}]%
        {Hansen2003}
\bibfield{author}{\bibinfo{person}{N. Hansen}, \bibinfo{person}{S.~D. M{\"{u}}ller}, {and} \bibinfo{person}{P. Koumoutsakos}.} \bibinfo{year}{2003}\natexlab{}.
\newblock \showarticletitle{{Reducing the Time Complexity of the Derandomized Evolution Strategy with Covariance Matrix Adaptation (CMA-ES)}}.
\newblock \bibinfo{journal}{\emph{Evolutionary Computation}} \bibinfo{volume}{11}, \bibinfo{number}{1} (\bibinfo{year}{2003}), \bibinfo{pages}{1--18}.
\newblock


\bibitem[\protect\citeauthoryear{Karshenas, Santana, Bielza, and Larra{\~{n}}aga}{Karshenas et~al\mbox{.}}{2012}]%
        {Karshenas2012}
\bibfield{author}{\bibinfo{person}{H. Karshenas}, \bibinfo{person}{R. Santana}, \bibinfo{person}{C. Bielza}, {and} \bibinfo{person}{P. Larra{\~{n}}aga}.} \bibinfo{year}{2012}\natexlab{}.
\newblock \showarticletitle{{Continuous Estimation of Distribution Algorithms Based on Factorized Gaussian Markov Networks}}.
\newblock In \bibinfo{booktitle}{\emph{Markov Networks in Evolutionary Computation}}. \bibinfo{publisher}{Springer}, \bibinfo{pages}{157--173}.
\newblock


\bibitem[\protect\citeauthoryear{Komarnicki, Przewozniczek, Kwasnicka, and Walkowiak}{Komarnicki et~al\mbox{.}}{2023}]%
        {Komarnicki2023}
\bibfield{author}{\bibinfo{person}{M.~M. Komarnicki}, \bibinfo{person}{M.~W. Przewozniczek}, \bibinfo{person}{H. Kwasnicka}, {and} \bibinfo{person}{K. Walkowiak}.} \bibinfo{year}{2023}\natexlab{}.
\newblock \showarticletitle{{Incremental Recursive Ranking Grouping for Large-Scale Global Optimization}}.
\newblock \bibinfo{journal}{\emph{IEEE Transactions on Evolutionary Computation}} \bibinfo{volume}{27}, \bibinfo{number}{5} (\bibinfo{year}{2023}), \bibinfo{pages}{1498--1513}.
\newblock


\bibitem[\protect\citeauthoryear{Olieman, Bouter, and Bosman}{Olieman et~al\mbox{.}}{2021}]%
        {Olieman2021}
\bibfield{author}{\bibinfo{person}{C. Olieman}, \bibinfo{person}{A. Bouter}, {and} \bibinfo{person}{P.~A.~N. Bosman}.} \bibinfo{year}{2021}\natexlab{}.
\newblock \showarticletitle{{Fitness-Based Linkage Learning in the Real-Valued Gene-Pool Optimal Mixing Evolutionary Algorithm}}.
\newblock \bibinfo{journal}{\emph{IEEE Transactions on Evolutionary Computation}} \bibinfo{volume}{25}, \bibinfo{number}{2} (\bibinfo{year}{2021}), \bibinfo{pages}{358--370}.
\newblock


\bibitem[\protect\citeauthoryear{Pelikan and Goldberg}{Pelikan and Goldberg}{2006}]%
        {Pelikan2005}
\bibfield{author}{\bibinfo{person}{M. Pelikan} {and} \bibinfo{person}{D.~E. Goldberg}.} \bibinfo{year}{2006}\natexlab{}.
\newblock \showarticletitle{{Hierarchical Bayesian Optimization Algorithm}}.
\newblock In \bibinfo{booktitle}{\emph{Studies in Computational Intelligence}}. Vol.~\bibinfo{volume}{33}. \bibinfo{publisher}{Springer}, \bibinfo{pages}{63--90}.
\newblock


\bibitem[\protect\citeauthoryear{Pelikan, Goldberg, and Cant{\'{u}}-Paz}{Pelikan et~al\mbox{.}}{1999}]%
        {Pelikan1999}
\bibfield{author}{\bibinfo{person}{M. Pelikan}, \bibinfo{person}{D.~E. Goldberg}, {and} \bibinfo{person}{E. Cant{\'{u}}-Paz}.} \bibinfo{year}{1999}\natexlab{}.
\newblock \showarticletitle{{BOA: The Bayesian Optimization Algorithm}}. In \bibinfo{booktitle}{\emph{Proceedings of the 1999 Conference on Genetic and Evolutionary Computation}}. \bibinfo{pages}{525--532}.
\newblock


\bibitem[\protect\citeauthoryear{Przewozniczek, Tin{\'{o}}s, Frej, and Komarnicki}{Przewozniczek et~al\mbox{.}}{2022}]%
        {Przewozniczek2022}
\bibfield{author}{\bibinfo{person}{M.~W. Przewozniczek}, \bibinfo{person}{R. Tin{\'{o}}s}, \bibinfo{person}{B. Frej}, {and} \bibinfo{person}{M.~M. Komarnicki}.} \bibinfo{year}{2022}\natexlab{}.
\newblock \showarticletitle{{On turning black - into dark gray-optimization with the direct empirical linkage discovery and partition crossover}}. In \bibinfo{booktitle}{\emph{Proceedings of the 2022 Genetic and Evolutionary Computation Conference}}. \bibinfo{pages}{269--277}.
\newblock


\bibitem[\protect\citeauthoryear{Quevedo De~Carvalho, Tinos, Whitley, and Sipoli~Sanches}{Quevedo De~Carvalho et~al\mbox{.}}{2019}]%
        {Carvalho2019}
\bibfield{author}{\bibinfo{person}{O. Quevedo De~Carvalho}, \bibinfo{person}{R. Tinos}, \bibinfo{person}{D. Whitley}, {and} \bibinfo{person}{D. Sipoli~Sanches}.} \bibinfo{year}{2019}\natexlab{}.
\newblock \showarticletitle{{A new method for identification of recombining components in the generalized partition crossover}}. In \bibinfo{booktitle}{\emph{Proceedings of the 2019 Brazilian Conference on Intelligent Systems}}. \bibinfo{pages}{36--41}.
\newblock


\bibitem[\protect\citeauthoryear{Shakya, Brownlee, McCall, Fournier, and Owusu}{Shakya et~al\mbox{.}}{2010}]%
        {Shakya2010}
\bibfield{author}{\bibinfo{person}{S. Shakya}, \bibinfo{person}{A. Brownlee}, \bibinfo{person}{J. McCall}, \bibinfo{person}{F. Fournier}, {and} \bibinfo{person}{G. Owusu}.} \bibinfo{year}{2010}\natexlab{}.
\newblock \showarticletitle{{DEUM - A Fully Multivariate EDA Based on Markov Networks}}.
\newblock In \bibinfo{booktitle}{\emph{Exploitation of Linkage Learning in Evolutionary Algorithms}}. Vol.~\bibinfo{volume}{3}. \bibinfo{publisher}{Springer}, \bibinfo{pages}{71--93}.
\newblock


\bibitem[\protect\citeauthoryear{Shakya and Santana}{Shakya and Santana}{2008}]%
        {Shakya2008}
\bibfield{author}{\bibinfo{person}{S. Shakya} {and} \bibinfo{person}{R. Santana}.} \bibinfo{year}{2008}\natexlab{}.
\newblock \showarticletitle{{An EDA Based on Local Markov Property and Gibbs Sampling}}. In \bibinfo{booktitle}{\emph{Proceedings of the 2008 Conference on Genetic and Evolutionary Computation}}. \bibinfo{pages}{475--476}.
\newblock


\bibitem[\protect\citeauthoryear{Shakya and Santana}{Shakya and Santana}{2012}]%
        {Shakya2012}
\bibfield{author}{\bibinfo{person}{S. Shakya} {and} \bibinfo{person}{R. Santana}.} \bibinfo{year}{2012}\natexlab{}.
\newblock \showarticletitle{{A Review of Estimation of Distribution Algorithms and Markov Networks}}.
\newblock In \bibinfo{booktitle}{\emph{Markov Networks in Evolutionary Computation}}. \bibinfo{publisher}{Springer}, \bibinfo{pages}{21--37}.
\newblock


\bibitem[\protect\citeauthoryear{Thierens and Bosman}{Thierens and Bosman}{2011}]%
        {Thierens2011}
\bibfield{author}{\bibinfo{person}{D. Thierens} {and} \bibinfo{person}{P.~A.~N. Bosman}.} \bibinfo{year}{2011}\natexlab{}.
\newblock \showarticletitle{{Optimal mixing evolutionary algorithms}}. In \bibinfo{booktitle}{\emph{Proceedings of the 2011 Genetic and Evolutionary Computation Conference}}. \bibinfo{pages}{617--624}.
\newblock


\bibitem[\protect\citeauthoryear{Tin{\'{o}}s, Whitley, and Chicano}{Tin{\'{o}}s et~al\mbox{.}}{2015}]%
        {Tinos2015}
\bibfield{author}{\bibinfo{person}{R. Tin{\'{o}}s}, \bibinfo{person}{D. Whitley}, {and} \bibinfo{person}{F. Chicano}.} \bibinfo{year}{2015}\natexlab{}.
\newblock \showarticletitle{{Partition crossover for Pseudo-Boolean optimization}}. In \bibinfo{booktitle}{\emph{Proceedings of the 2015 ACM Conference on Foundations of Genetic Algorithms XIII}}. \bibinfo{publisher}{Association for Computing Machinery, Inc}, \bibinfo{pages}{137--149}.
\newblock


\end{thebibliography}

\end{document}